\documentclass[10pt,twocolumn,letterpaper]{article}

\usepackage{iccv}
\usepackage{times}
\usepackage{epsfig}
\usepackage{graphicx}
\usepackage{amsmath}
\usepackage{amssymb}
\usepackage{enumitem}

\usepackage{multirow}
\usepackage{array}
\usepackage{dsfont}
\usepackage{boldline}
\usepackage{dblfloatfix}
\usepackage[titletoc,title]{appendix}

\usepackage[pagebackref=true,breaklinks=true,letterpaper=true,colorlinks,bookmarks=false]{hyperref}

\iccvfinalcopy 
\def\iccvPaperID{382}

\newcommand{\bst}[1]{\textcolor{red}{ \textbf{#1} }}
\newcommand{\scd}[1]{\textcolor{blue}{ \underline{#1} }}

\ificcvfinal\fi
\begin{document}

\title{LabelBank: Revisiting Global Perspectives for Semantic Segmentation}

%

\author{Hexiang Hu\thanks{equal contribution.}\\
University of Southern California\\
Los Angeles, CA\\
{\tt\small hexiang.frank.hu@gmail.com}
\and
Zhiwei Deng\footnotemark[1]\\
Simon Fraser University\\
Burnaby, BC, Canada\\
{\tt\small zhiweid@sfu.ca}
\and
Guang-Tong Zhou\\
Oracle Labs\\
Vancouver, BC, Canada\\
{\tt\small zhouguangtong@gmail.com}
\and
Fei Sha\\
University of Southern California\\
Los Angeles, CA\\
{\tt\small feisha@usc.edu}
\and
Greg Mori\\
Simon Fraser University\\
Burnaby, BC, Canada\\
{\tt\small mori@cs.sfu.ca}
}

\maketitle

\begin{abstract}
Semantic segmentation requires a detailed labeling of image pixels by object category. Information derived from local image patches is necessary to describe the detailed shape of individual objects. However, this information is ambiguous and can result in noisy labels. Global inference of image content can instead capture the general semantic concepts present. We advocate that holistic inference of image concepts provides valuable information for detailed pixel labeling. We propose a generic framework to leverage holistic information in the form of a \textit{LabelBank} for pixel-level segmentation. 

We show the ability of our framework to improve semantic segmentation performance in a variety of settings. We learn models for extracting a holistic \textit{LabelBank} from visual cues, attributes, and/or textual descriptions. We demonstrate improvements in semantic segmentation accuracy on standard datasets across a range of state-of-the-art segmentation architectures and holistic inference approaches.

\end{abstract}


\vspace{-0.1in}
\section{Introduction}
\label{sec:intro}
\vspace{-0.05in}

Great progress has been made in visual recognition.  In tasks ranging from image classification to object detection, algorithms rival human performance in certain conditions.  Semantic segmentation, labeling each pixel to depict semantic elements by detailed shapes and contours, is arguably a requisite element of full visual understanding of a scene.  For applications such as robotics, autonomous driving, or other scene understanding endeavours, accurate delineation of object contours is necessary for success.

\begin{figure}[t]
\vspace{-0.1in}
\begin{center}
\includegraphics[width=0.475\textwidth]{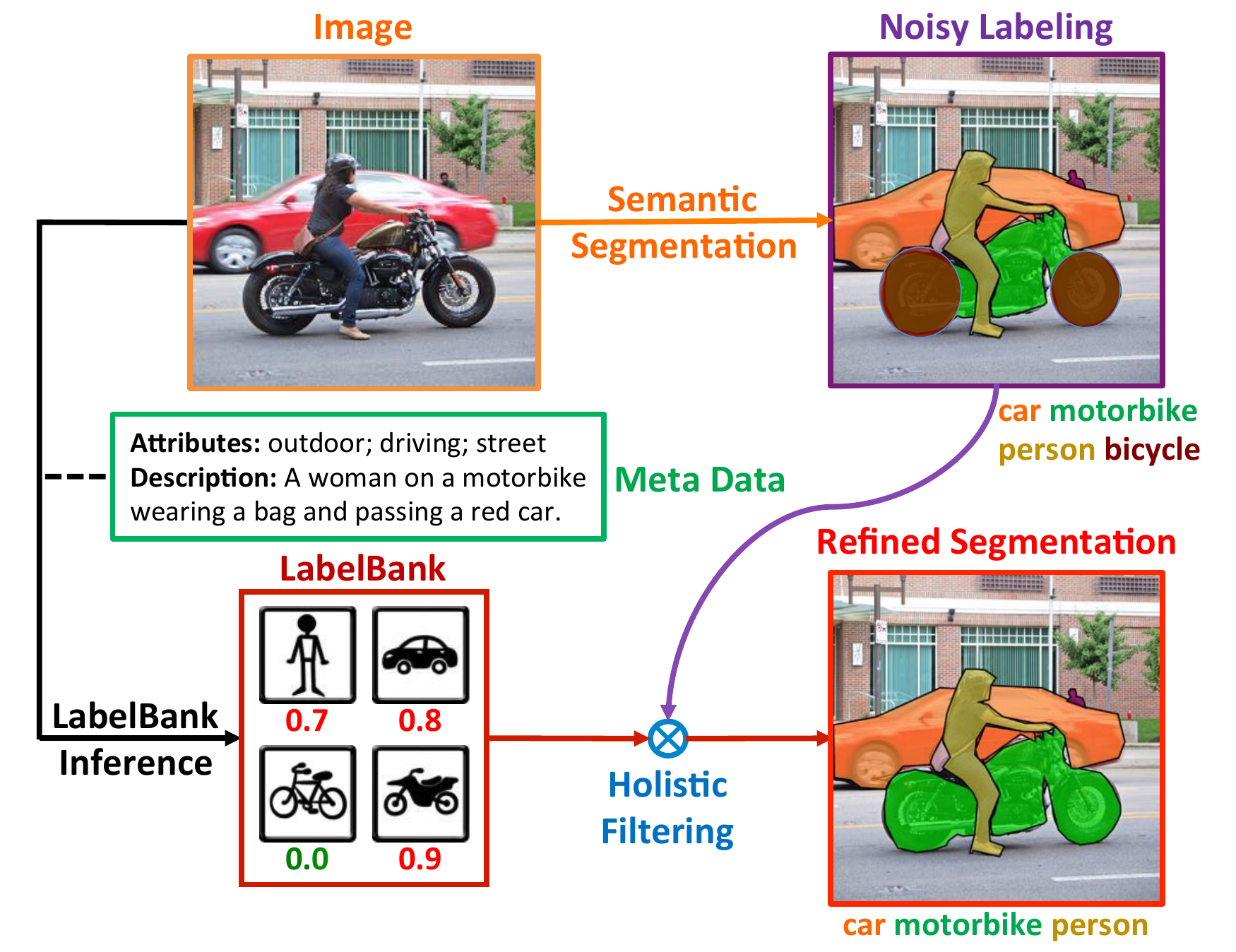}
\end{center}
\caption{An example showing the usage of holistic LabelBank in semantic segmentation. We obtain a LabelBank representation of an image from visual appearance and / or meta data, and leverage it to filter out false-positive pixel predictions to improve semantic segmentation. In this example, the false-positive predictions of bicycle are removed after holistic filtering since bicycle is not suggested as a likely label in the LabelBank.}
\label{fig:intro}
\vspace{-0.1in}
\end{figure}

However, detailed semantic segmentation is challenging -- there exists significant ambiguity in fine-scale image patches that can result in noisy semantic segmentation outputs.  The main focus of this paper is utilizing holistic information to filter noisy low-level semantic segmentation (see Figure~\ref{fig:intro}).  We term this holistic representation \textbf{LabelBank}, specifically defined as a continuous vector of confidences of which objects are likely to be present in an image.

This holistic LabelBank can be derived from a variety of sources.  Akin to image classification, one could directly infer object content from global visual features describing an image (c.f.\ the seminal ObjectBank work~\cite{li_nips10}).  Further, this information can be extracted more generally -- in many scenarios we have additional meta data such as sentences describing an image or tag-style labels regarding image attributes.  We demonstrate that our framework can be utilized in a range of settings, from purely visual information (no additional data beyond the image pixels) to situations where additional meta data are present.

We instantiate our ideas within a single framework for inferring the LabelBank representation and utilizing it for semantic segmentation.  This framework can be used in common with a variety of state-of-the-art semantic segmentation networks.
State-of-the-art methods for semantic segmentation leverage the successes of Convolutional Neural Networks (CNNs)~\cite{lecun_pieee98}. CNNs have transformed the field of image classification, especially since the development of AlexNet~\cite{alex_nips12}. There have been many follow-up CNN architectures to further boost image classification, including VGGNet~\cite{simonyan_iclr15}, Google Inception~\cite{szegedy_cvpr15}, ResNet~\cite{he_cvpr16}, \textit{etc}. Semantic segmentation utilizes these network structures, combined with dense output structures to label image pixels by semantic categories. A representative work is the Fully Convolutional Network (FCN)~\cite{shelhamer_pami16} that leverages skip features of CNNs to produce a detailed pixel labeling. Another example is the DeepLab~\cite{chen_arxiv16} framework, which augments FCN with dilated convolution~\cite{yu_iclr16}, atrous spatial pyramid pooling and Conditional Random Fields (CRFs), and obtains state-of-art semantic segmentation performance.

Common among these previous semantic segmentation methods is a focus on (layers of) low-level pixel analysis leading to semantic segmentation.  State-of-the-art techniques combine this with graphical model-style techniques (CRF) and pooling structures to obtain high accuracy.  The role of these additional components is to smooth out the noisy pixel labelings that result from the direct CNN analysis.  As a complementary approach, we advocate for a holistic inference of LabelBank that globally suggests category labels that are likely to be present in the image.

To make use of the LabelBank representation in semantic segmentation, we leverage it to filter out false-positive pixel predictions -- if the holistic information in the LabelBank suggests a semantic category is unlikely to be present, then pixels should be unlikely to be predicted as that label. We utilize these observations to propose a framework that unifies a LabelBank inference process and a detailed semantic segmentation process via a holistic filtering process. Our framework is generic and flexible enough to leverage different data sources / architectures in LabelBank inference and can integrate state-of-the-art semantic segmentation networks. For example, the LabelBank can be derived from cues ranging from image appearance to attributes to textual descriptions. The semantic segmentation process can be implemented using state-of-the-art approaches, such as FCN~\cite{shelhamer_pami16} and DilatedNet~\cite{yu_iclr16,chen_arxiv16}. Finally, our holistic filtering leverages the information in the LabelBank to guide segmentation by refining noisy pixel predictions.


\noindent{\bf Contribution.} We summarize our main contributions as:
\vspace{-0.05in}
\begin{itemize}[leftmargin=*]
\setlength{\itemsep}{0pt}
\setlength{\parskip}{0pt}
\item First, we develop LabelBank to guide semantic segmentation, where LabelBank is a holistic representation of image content that can be derived from various sources.
\item Second, we construct holistic filtering that enables us to filter out false-positive pixel predictions under the guidance of LabelBank.
\item Third, we propose a neural network framework for semantic segmentation. We implement approaches for inferring LabelBank and conducting semantic segmentation, which facilitate the flow of global image information to pixel segmentation. Our framework is general, and could be incorporated into a variety of CNN-based semantic segmentation architectures.
\item Finally, we evaluate our proposed framework on standard semantic segmentation datasets: PASCAL-Context~\cite{mottaghi_cvpr14}, ADE20K~\cite{zhou_arxiv16}, COCO-Stuff~\cite{caesar_arxiv16}, NYUDv2~\cite{silberman_eccv12} and SIFT-Flow~\cite{liu_pami11}. Experimental results show that the proposed LabelBank-based framework can be used to improve a variety of state-of-the-art semantic segmentation approaches.
\end{itemize}

\vspace{-0.1in}
\section{Related Work}
\label{sec:related}
\vspace{-0.05in}

\begin{figure*}[tp]
\vspace{-0.1in}
\centering
\includegraphics[width=1\textwidth]{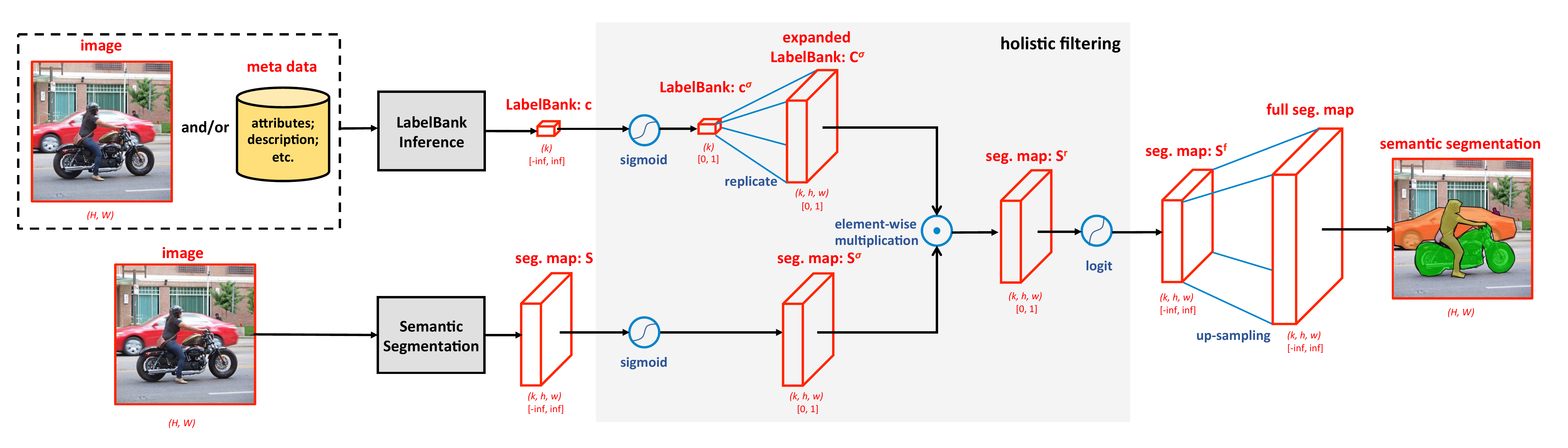}
\caption{Our semantic segmentation framework. Please refer to the text for a detailed description. The notation in curly brackets and square brackets respectively depict the size and the value domain of the data.}
\label{fig:framework:pipeline}
\vspace{-0.05in}
\end{figure*}


\noindent\textbf{Semantic segmentation}. The success of CNNs in object recognition has led to renewed attention on semantic segmentation. A representative work is the Fully Convolutional Network (FCN)~\cite{shelhamer_pami16} that uses skip features of CNNs for detailed pixel labeling. FCN combines multi-level feature descriptors to leverage coarse-to-fine local pixel information. In a recent advance, the atrous convolution is introduced by Chen~\textit{et al.}~\cite{chen_arxiv16} as a technique to retain a large field of view while keeping fewer trainable weights in semantic segmentation networks. The same method termed as dilated convolution was also pursued by Yu and Koltun~\cite{yu_iclr16} by a cascading series of dilated convolution layers.

Another line of work pushes on refining the detailed shapes and contours of semantic segmentation. A fully connected CRF is proposed by Kr{\"{a}}henb{\"{u}}hl and Koltun~\cite{krahenbuhl_nips12} as an efficient dense pixel modeling method. The fully connected pairwise CRF is adopted by Chen~\textit{et al.}~\cite{chen_arxiv16} and Zheng~\textit{et al.}~\cite{zheng_iccv15} on top of FCNs as a further refinement. These methods have achieved considerable improvement on the semantic segmentation task. Different from the above-mentioned methods, our work leverages holistic LabelBank to filter out false-positive pixel predictions.

\noindent\textbf{Global-local information fusion}. It has been shown that visual understanding benefits from exploiting and leveraging information of varying granularity. Deng~\textit{et al.}~\cite{deng_eccv14} modeled hierarchical and exclusive relations among semantic categories. Jain~\textit{et al.}~\cite{jain_cvpr16} developed recurrent neural network structures for spatio-temporal inference. Hu~\textit{et al.}~\cite{hu_cvpr16} proposed a neural graph inference model to propagate information among multiple levels of visual classes, including coarse labels, fine-grained categories and attributes. Amer~\textit{et al.}~\cite{amer_eccv12} adopted the and-or graph structure to reason about human activities at multiple levels of granularity. Gkioxari~\textit{et al.}~\cite{gkioxari_iccv15} exploited contextual cues to improve action recognition.

In the realm of semantic segmentation, He~\textit{et al.}~\cite{he_cvpr04} proposed to use multi-scale CRFs to capture features at various image resolutions for semantic segmentation. Approaches for modeling object instances and their segmentations have also been developed~\cite{kumar_cvpr05,wu_cvpr07}. Another example of this line improves instance-aware semantic segmentation with object detection and classification~\cite{dai_cvpr15,dai_cvpr16}. In contrast, we exploit holistic LabelBank for semantic segmentation, and do not require instance-level annotations.

\section{Framework}
\label{sec:framework}
\vspace{-0.05in}

We now present our framework that leverages a holistic LabelBank in semantic segmentation. Recall that the LabelBank is defined as a continuous vector of confidences for the presence of semantic object categories in an image.

Figure~\ref{fig:framework:pipeline} provides an overview of our framework. In detail, it is composed of three components. First, we have a holistic inference process that takes varied information sources to reason about the LabelBank representation of an image. Second, we have a detailed semantic segmentation process to conduct preliminary semantic segmentation on the image to generate a segmentation map. Finally, we have a holistic filtering process that leverages the inferred LabelBank to filter out false-positive pixel predictions in the preliminary semantic segmentation results.

Our framework is generic and can flexibly incorporate various data sources / architectures in LabelBank inference, leveraging different semantic segmentation networks. For example, the LabelBank can be derived from a variety of data sources, ranging from purely visual appearance to the cases where additional meta data are available, such as sentences describing image content and tag-like labels on image attributes. The LabelBank inference architecture also varies depending on the available data sources. Furthermore, our semantic segmentation process is also generic, and can be implemented with state-of-the-art CNN-based network architectures like FCN~\cite{shelhamer_pami16} or DilatedNet~\cite{yu_iclr16,chen_arxiv16}. In our implementation, we slightly modified FCN and DilatedNet for improved performance, by replacing the linear pixel classifier with a non-linear two-layer CNN. Due to space limitations, we present the details in the appendix.

We report our experimental results in Section~\ref{sec:experiment} to verify the generalizability and effectiveness of our framework. In what follows, we first describe our holistic filtering process in Section~\ref{sec:framework:filter}, and then present our exemplar implementations of LabelBank inference in Section~\ref{sec:framework:inference}.

\subsection{Holistic Filtering}
\label{sec:framework:filter}
\vspace{-0.05in}

Holistic filtering is a key component in our framework. It uses the LabelBank representation derived from the holistic inference process to actively filter out false-positive pixel predictions in the segmentation map generated by the semantic segmentation process. Note that a segmentation map is typically organized as a matrix of confidences for assigning each semantic category label on each image pixel. Our idea is to use the LabelBank to recommend labels for pixel predictions -- if LabelBank suggests a semantic label is unlikely to be present, then pixels should be unlikely to be predicted as that label as well.

To implement the idea, we weight the segmentation map predictions on each pixel by the LabelBank confidences. The weighting is done by a multiplication of both LabelBank and segmentation confidences transformed in a sigmoidal space. In the sigmoidal space, unlikely labels tend to receive low confidences close to 0, and likely labels tend to have high confidences close to 1. Therefore, the final confidence after the multiplication is high only if both the LabelBank and segmentation confidences are high. The final confidence is low whenever either the LabelBank confidence or the segmentation confidence is low. The detailed holistic filtering process is illustrated in Figure~\ref{fig:framework:pipeline}.

Formally, we denote the LabelBank representation of an image by a vector $\mathbf{c} \in \mathbb{R}^{k \times 1 \times 1}$, where $k$ is the total number of semantic labels of interest. Each element of $\mathbf{c}$ indicates the confidence of observing the corresponding semantic category in the image. We also denote $\mathbf{S} \in \mathbb{R}^{k \times h \times w}$ as the segmentation map generated by the semantic segmentation process, where each element of $\mathbf{S}$ stores the confidence of observing a semantic category at the corresponding image location. Note that the segmentation map size $h \times w$ is typically smaller than the original image size $H \times W$, due to the pooling or down-sampling operations employed by most semantic segmentation networks (\textit{e.g.}, FCN~\cite{shelhamer_pami16} and DilatedNet~\cite{yu_iclr16,chen_arxiv16}).

We first map the LabelBank $\mathbf{c}$ and the segmentation map $\mathbf{S}$ to a sigmoidal space via:
\vspace{-0.05in}
\begin{equation}
\label{eqn:framework:sigmoid}
\mathbf{c}^{\sigma} = \big[ \sigma(\mathbf{c}_{l}) \big]_{l=1}^{l=k}, \ \ \ \ \mathbf{S}^{\sigma} = \big[ \sigma(\mathbf{S}_{l,i,j}) \big]_{l=1,i=1,j=1}^{l=k,i=h,j=w},
\vspace{-0.05in}
\end{equation}
where $\sigma(x) = \frac{1}{1 + e^{-x}}$ is the sigmoid function.

To weight each location of the segmentation map by the LabelBank, we replicate $\mathbf{c}^{\sigma}$ to obtain an expanded LabelBank $\mathbf{C}^{\sigma} \in \mathbb{R}^{k \times h \times w}$ of the same size as the segmentation map. Note that the confidence vector of the expanded LabelBank at each location is a copy of the LabelBank, \textit{i.e.},
\vspace{-0.05in}
\begin{equation}
\label{eqn:framework:replicate}
\mathbf{C}^{\sigma}_{:,i,j} = \mathbf{c}^{\sigma}, \ \ \forall 1 \leq i \leq h, 1 \leq j \leq w.
\vspace{-0.05in}
\end{equation}
We then conduct element-wise multiplication of the expanded LabelBank and the segmentation map to filter out false-positive predictions in the segmentation map. The refined segmentation map is computed as:
\vspace{-0.05in}
\begin{equation}
\label{eqn:framework:multiply}
\mathbf{S}^{r} = \mathbf{C}^{\sigma} \odot \mathbf{S}^{\sigma}.
\vspace{-0.05in}
\end{equation}
In $\mathbf{S}^{r}$, a location receives a high confidence on a label only if both the LabelBank and the original pixel prediction are highly confident of predicting that label.

Finally, we apply a logit function (\textit{i.e.}, the inverse sigmoid function) on each element of the refined segmentation map, so that the final confidences share the same value domain as the original segmentation map $\mathbf{S}$. Formally, the LabelBank filtered segmentation map is derived as:
\vspace{-0.05in}
\begin{equation}
\label{eqn:framework:logit}
\mathbf{S}^{f} = \big[ \ell(\mathbf{S}^{r}_{l,i,j}) \big]_{l=1,i=1,j=1}^{l=k,i=h,j=w},
\vspace{-0.05in}
\end{equation}
where $\ell(x) = \log(\frac{x}{1 - x})$ is the logit function. Note that all operations in our holistic filtering process are differentiable for gradient back-propagation to enable end-to-end training.

As $\mathbf{S^{f}}$ is typically sized smaller than the original image, we apply an up-sampling operation to generate a full semantic segmentation map. The up-sampling is simply done by bi-linear interpolation (following~\cite{chen_arxiv16}) to increase the resolution to the original image size. We could also switch the order of up-sampling and holistic filtering in the pipeline to leverage LabelBank to refine the full segmentation map. This results in slightly better empirical performance, but increases the computational cost significantly. We keep up-sampling after holistic filtering for the sake of efficiency.

\subsection{LabelBank Inference}
\label{sec:framework:inference}

The LabelBank can be inferred from a variety of data sources. A straightforward way is to use the image itself, \textit{i.e.}, visual appearance. We present two exemplar visual inference architectures in Sections~\ref{sec:framework:inference:spp} and \ref{sec:framework:inference:dc}, but other architectures are possible and can be easily adopted in our framework. Moreover, the LabelBank can also be derived from commonly available image meta data, such as tag-based labels on image attributes or sentences describing the image content. We implement sample architectures in Section~\ref{sec:framework:inference:mlp} for this purpose. Furthermore, we have also experimented with combined visual appearance and meta data for a more accurate LabelBank representation.



To design the architectures, we would like to emphasize the following two principles. First, we prefer a process that takes an image and / or its meta data as input and produces the LabelBank representation as output, where the gradient can be back-propagated through for end-to-end training. Second, the architectures should not require any additional supervision beyond the readily available pixel labels in training data. The pixel labels have already depicted the semantic objects in the images, and we should make use of them in LabelBank inference.


\begin{figure}[tp]
\vspace{-0.1in}
\centering
\includegraphics[width=0.475\textwidth]{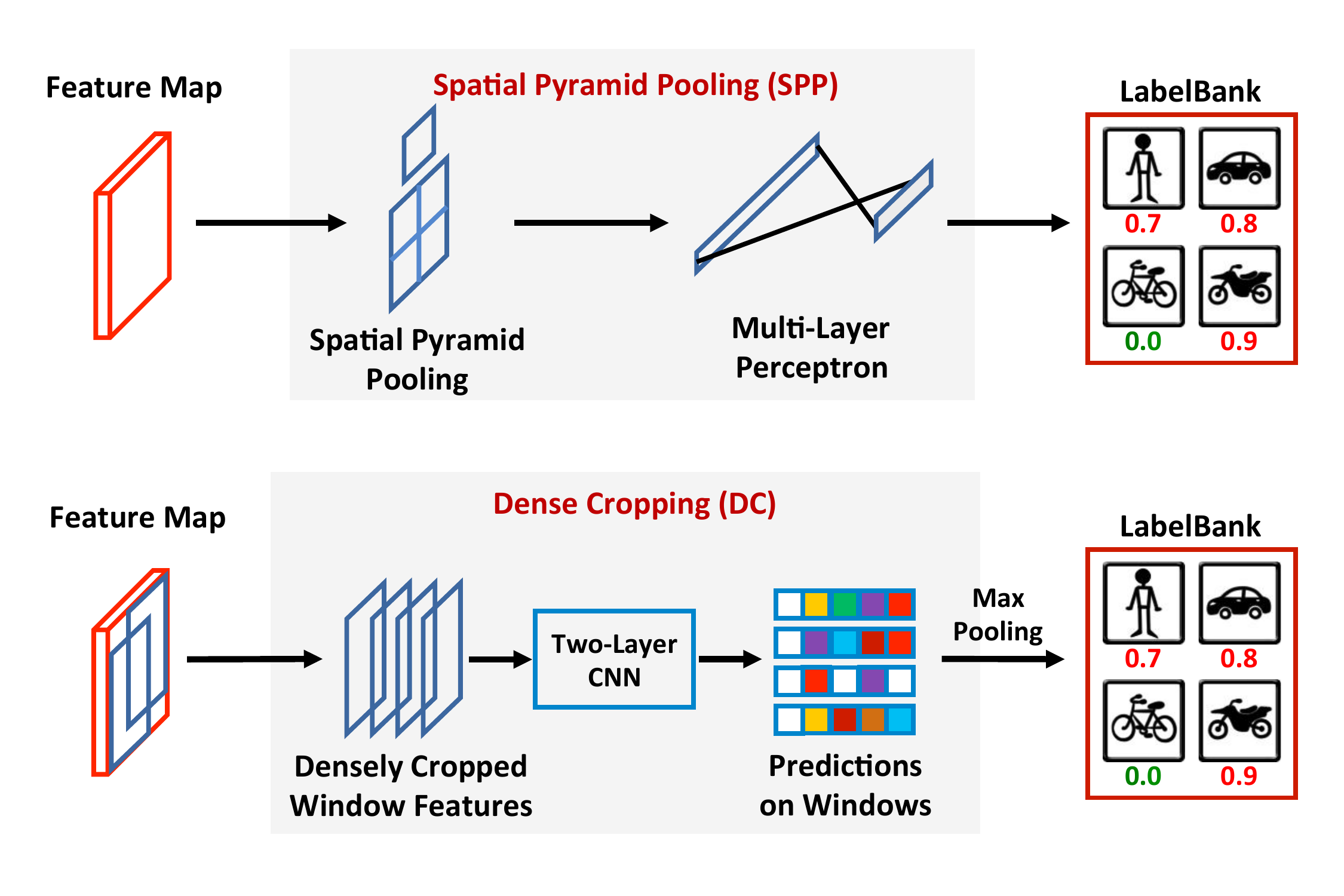}
\caption{The SPP and DC architectures for LabelBank inference.}
\label{fig:framework:inference}
\vspace{-0.05in}
\end{figure}

\vspace{-0.1in}
\subsubsection{SPP for Visual Appearance}
\label{sec:framework:inference:spp}
\vspace{-0.05in}

Following the design principles, we first implement a spatial pyramid pooling (SPP) based architecture for visual inference. This is illustrated in the top part of Figure~\ref{fig:framework:inference}.

The SPP architecture is motivated by the success of spatial pyramid pooling for image recognition~\cite{he_eccv14}. It first employs a feature network (\textit{i.e.}, low-level layers of convolution and pooling) to extract a feature map on an image. Then it applies spatial pyramid pooling on the feature map, followed by a multi-layer perception to predict the LabelBank.


For training purposes, we obtain ground-truth LabelBank from the ground-truth pixel labels -- an image is labeled by a semantic category if there is at least one pixel labeled with that category. We adopt a sigmoid activation layer on the output of the multi-layer perception, and evaluate a categorical cross-entropy loss.

\vspace{-0.1in}
\subsubsection{DC for Visual Appearance}
\label{sec:framework:inference:dc}
\vspace{-0.05in}

We further develop a dense cropping (DC) based architecture for visual inference. This is shown in the bottom part of Figure~\ref{fig:framework:inference}. The DC architecture is motivated by the fact that a semantic object rarely takes up the entire image, but instead a portion of an image. Thus, we conduct location-aware predictions that inspect densely cropped image windows.

For implementation, we first apply the same feature network as the SPP architecture. Then we densely crop the feature map to obtain features on image windows for location-aware predictions. We implement a two-layer CNN structure as the prediction model (details in the appendix). Each location-aware prediction results in a $k$-dimensional vector that describes the confidences for each semantic category to appear in the corresponding image window. Finally, we compose the LabelBank representation by a max-pooling over the location-aware predictions.

We impose a loss on the location-aware predictions. Specifically, the ground-truth labels on each image window are derived from the ground-truth pixel labels -- an image window is labeled by a semantic category as long as the image window has at least one pixel labeled with that category. We first apply a sigmoid activation layer on the output of the location-aware predictions, and then evaluate a categorical cross-entropy loss.


\vspace{-0.1in}
\subsubsection{OHE and W2V Embedding for Meta Data}
\label{sec:framework:inference:mlp}
\vspace{-0.05in}

We also propose an embedding based architecture to infer LabelBank from meta data. This architecture first embeds meta data in a feature space. Specifically, depending on the type of the meta data, we could apply one-hot encoding for attributes (OHE architecture), or word2vec-style representation~\cite{mikolov_nips13,pennington_emnlp14} for sentence descriptions (W2V architecture). With the embedded features, we then apply a multi-layer perception to predict the LabelBank of an image.

We train by imposing a loss on the final LabelBank predictions. The ground-truth LabelBank and loss are computed the same way as done in our SPP architecture.

\vspace{-0.1in}
\subsubsection{Training}
\label{sec:framework:inference:train}
\vspace{-0.05in}

Our framework is end-to-end trainable, because gradient back-propagation is enabled in the above LabelBank inference processes, preliminary semantic segmentation (FCN and DilatedNet), as well as holistic filtering (Section~\ref{sec:framework:filter}).

We jointly optimize both LabelBank inference loss and semantic segmentation loss in training. We have described the LabelBank inference losses in the above subsections. The segmentation loss is enforced on the full segmentation map obtained from holistic filtering. Following the standard setting of semantic segmentation~\cite{shelhamer_pami16,chen_arxiv16}, we first apply a softmax activation layer to the full segmentation map, and then compute a categorical cross-entropy loss for the segmentation task. The balance between the two losses is controlled by a constant multiplier so that both losses have similar order of magnitude. To further clarify our implementation details, we will publicly release our code.


\vspace{-0.1in}
\section{Experiments}
\label{sec:experiment}
\vspace{-0.05in}

In this section, we conduct experiments to verify the effectiveness and generalizability of our framework. We examine a variety of settings for LabelBank inference and semantic segmentation. We present the experiments from two perspectives. First, we study the variants of LabelBank inference in Section~\ref{sec:experiment:labelbank}, using different data sources and architectures. Second, we study the variants in our semantic segmentation process in Section~\ref{sec:experiment:segmentation}, using FCN and DilatedNet. Before diving into the details, we first describe the common experimental settings in Section~\ref{sec:experiment:setting}.


\subsection{Settings}
\label{sec:experiment:setting}
\vspace{-0.05in}

\noindent\textbf{Datasets.} We evaluate our framework on five benchmark datasets: PASCAL-Context~\cite{mottaghi_cvpr14}, ADE20K~\cite{zhou_arxiv16}, COCO-Stuff~\cite{caesar_arxiv16}, NYUDv2~\cite{silberman_eccv12}, and SIFT-Flow~\cite{liu_pami11}. All these datasets contain abundant contextual labels on pixels, and thus are challenging for semantic segmentation. Table~\ref{tab:experiment:dataset} summarizes the five datasets. The appendix elaborates on the details of these datasets.

\begin{table}[htp]
\vspace{-0.05in}
\small
\tabcolsep 4pt
\centering
\begin{tabular}{ c|c|c }
\small{Dataset}& \small{\# of classes} & \small{\# of train/val/test images} \\
\hlineB{4}
{Pascal-Context~\cite{mottaghi_cvpr14}} & 59 + bg & 4,998/5,105/- \\ 
{ADE20K~\cite{zhou_arxiv16}} & 150 + bg & 20,210/2,000/5,000 \\ 
{COCO-Stuff~\cite{caesar_arxiv16}} & 171 + bg & 9,000/1,000/- \\ 
{NYUDv2~\cite{silberman_eccv12}} & 40 & 795/654/-\\ 
{SIFT-Flow~\cite{liu_pami11}} & 33 & 2488/200/-\\ 
\hline
\end{tabular}
\caption{A summary of our experimental datasets.}
\label{tab:experiment:dataset}
\vspace{-0.05in}
\end{table}

\noindent\textbf{Baselines.} A direct baseline is to use the semantic segmentation process only, and ignore LabelBank inference and holistic filtering. As mentioned above, we modified FCN~\cite{shelhamer_pami16} and DilatedNet~\cite{yu_iclr16,chen_arxiv16} to be our segmentation networks, and refer them to FCN$^{+}$ and DilatedNet$^{+}$.

\noindent\textbf{Evaluation Protocols.} Following~\cite{shelhamer_pami16}, we evaluate four common performance metrics for semantic segmentation. The metrics are variations on pixel accuracy and region intersection over union (IU). Specifically, we denote by $n_{ij}$ the number of pixels of category $i$ predicted to belong to category $j$, $t_{i} = \sum_{j} n_{ij}$ the total number of pixels of category $i$, and $k$ the total number of categories. We evaluate:
\vspace{-0.2in}
\begin{itemize}[leftmargin=*]
\setlength{\itemsep}{1pt}
\setlength{\parskip}{1pt}
\setlength{\parsep}{1pt}
\item Pixel Accuracy (pAcc): $\sum_{i} n_{ii} / \sum_{i} t_{i}$
\item Mean Accuracy (mAcc): $(1/k) \sum_{i} n_{ii}/t_{i}$
\item Mean IU (mIU): $(1/k) \sum_{i} n_{ii} / \big(t_{i} + \sum_{j} n_{ji} - n_{ii}\big)$
\item Frequency Weighted IU (fwIU): \\
$(\sum_{k} t_{k})^{-1} \sum_{i} t_{i} n_{ii} / \big(t_{i} + \sum_{j} n_{ji} - n_{ii}\big)$
\end{itemize}
\vspace{-0.05in}
In all the result tables, we highlight the best results in red and boldfaced, and the second best in blue and underline.

\noindent\textbf{Fine Print.} We provide the detailed data augmentation, feature networks, training strategies and qualitative visualizations in the appendix.

\vspace{-0.05in}
\subsection{Study on the LabelBank Inference Process}
\label{sec:experiment:labelbank}
\vspace{-0.05in}

In this section, we evaluate variations on LabelBank inference, with different data sources and architectures. Specifically, we have SPP and DC for visual appearance, OHE for attributes, and W2V for textual descriptions. Here we fix our semantic segmentation network as FCN$^{+}$, and show that our framework is effective and generic with respect to various LabelBank inference processes.

\vspace{-0.05in}
\subsubsection{Inference from Visual Appearance}
\label{sec:experiment:labelbank:visual}
\vspace{-0.05in}

We first instantiate our framework to infer LabelBank from image visual appearance, using SPP and DC. This setting requires no additional meta data, and is directly comparable with existing semantic segmentation approaches. We show that our framework achieves favorable performance (the results on NYUDv2 and SIFT-Flow are deferred to the appendix due to space limit).

\begin{table}[tp!]
\vspace{-0.1in}
\small
\tabcolsep 4pt
\centering
\begin{tabular}{ c|cccc }
 & pAcc & mAcc & mIU & fwIU \\
\hlineB{4}
{CFM~\cite{dai_cvpr15}}                         & -           & -           & 34.4            & -           \\ 
{FCN-8s~\cite{shelhamer_pami16}}                & 67.45       & 52.31       & 39.12           & 53.03       \\ 
{CRF-RNN~\cite{zheng_iccv15}}                   & -           & -           & 39.3            & -           \\ 
{DeepLab~\cite{chen_arxiv16}}                   & -           & -           & 39.6            & -           \\ 
{ParseNet~\cite{liu_iclr16}}                    & -           & -           & 40.4            & -           \\ 
{BoxSup~\cite{dai_iccv15}}                      & -           & -           & 40.5            & -           \\ 
{HO-CRF~\cite{arnab_eccv16}}                    & -           & -           & 41.3            & -           \\ 
{Context~\cite{lin_cvpr16}}                     & 71.5        & 53.9        & 43.3            & -           \\ 
{DeepLab + COCO~\cite{chen_arxiv16}}            & -           & -           & 44.7            & -           \\ 
{DeepLab + COCO + CRF~\cite{chen_arxiv16}}      & -           & -           & \scd{45.7}            & -           \\ 
\hline
{FCN$^{+}$}                          & 71.25 & 53.82 & 43.19 & 57.63 \\ 
{SPP + FCN$^{+}$}                    & \scd{72.35} & \scd{55.06} & 44.32 & \scd{58.95} \\
{DC + FCN$^{+}$}                     & \bst{73.52} & \bst{56.72} & \bst{45.77}  & \bst{60.05} \\
\hline
\end{tabular}
\caption{Semantic segmentation results on PASCAL-Context.}
\label{tab:experiment:pascal}
\vspace{-0.1in}
\end{table}


\noindent\textbf{PASCAL-Context.} We have compared our methods (SPP + FCN$^{+}$ and DC + FCN$^{+}$) with the baseline FCN$^{+}$, as well as existing methods in the literature. The results are reported in Table~\ref{tab:experiment:pascal}. It clearly shows that our methods obtain better performance over all the compared methods. This verifies the effectiveness of our framework. Table~\ref{tab:experiment:pascal} also validates the utility of LabelBank based holistic filtering -- it boosts FCN$^{+}$ substantially.

It is worth noting that DC+ FCN$^{+}$ outperforms the current state-of-the-art method, DeepLab + COCO + CRF~\cite{chen_arxiv16}. However, we did not use extra training data (\textit{e.g.} the COCO dataset) and domain adaptation to obtain a better model. We applied neither CRFs to smooth the segmentation results, nor multi-scale test to refine segmentation at various image granularities. Therefore, we predict that our performance could be further boosted using these techniques.



\noindent\textbf{ADE20K.} We follow the settings of~\cite{zhou_arxiv16} in our experiments. We train and test our frameworks on re-sized images of 384$\times$384 pixels. To evaluate the performance against the ground-truth annotations, we re-size the resultant segmentation from 384$\times$384 back to the original image size. This experiment enables us to directly compare with those reported in~\cite{zhou_arxiv16}. Note that this setting may lead to the loss of image details and object aspect ratios, thus yielding sub-optimal performance. We summarize the results in Table~\ref{tab:experiment:ade-384}. It shows that our DC + FCN$^{+}$ achieves the best performance in all four metrics, when compared with the baseline FCN$^{+}$ and existing approaches in literature.

\begin{table}[tp!]
\vspace{-0.1in}
\small
\tabcolsep 5.5pt
\centering
\begin{tabular}{ c|cccc }
 & pAcc & mAcc & mIU & fwIU \\
\hlineB{4}
{SegNet~\cite{zhou_arxiv16}}                & 71.00       & 31.14       & 21.64       & 53.84  \\ 
{SegNet Cascade~\cite{zhou_arxiv16}}        & 71.83       & 37.90       & 27.51       & 58.05  \\ 
{FCN-8s~\cite{zhou_arxiv16}}                & 71.32       & 40.32       & 29.39       & 57.33  \\ 
{DilatedNet~\cite{zhou_arxiv16}}            & 73.55       & 44.59       & 32.31       & 60.14  \\ 
{DilatedNet Cascade~\cite{zhou_arxiv16}}    & 74.52       & 45.38       & 34.90       & 61.08  \\ 
\hline
{FCN$^{+}$}                                 & 77.30       & 46.94       & 36.52       & 64.49 \\
{SPP + FCN$^{+}$}                           & 77.99       & 45.98       & 36.61       & 65.30 \\
{DC + FCN$^{+}$}                            & 78.03       & \scd{48.23} & 37.93       & 65.34 \\
\hline
{OHE + FCN$^{+}$}                           & \scd{79.62} & 47.66     & \scd{38.12} & \scd{67.30} \\
{OHE + DC + FCN$^{+}$}                      & \bst{82.26} & \bst{53.18} & \bst{43.46} & \bst{70.82} \\
\hline
\end{tabular}
\caption{Semantic segmentation results on the ADE20K validation set, using 384$\times$384 training and testing images.}
\label{tab:experiment:ade-384}
\vspace{-0.05in}
\end{table}

\begin{table}[tp!]
\small
\tabcolsep 7pt
\centering
\begin{tabular}{ c|cccc }
 & pAcc & mAcc & mIU & fwIU \\
\hlineB{4}
{FCN~\cite{caesar_arxiv16}}             & {52.0} & {34.0} & {22.7} & - \\
{DeepLab~\cite{caesar_arxiv16}}         & {57.8} & {38.1} & {26.9} & - \\
\hline
{FCN$^{+}$}                             & {62.40} & {43.10} & {30.77} & {47.98} \\
{SPP + FCN$^{+}$}                       & {63.37} & {40.48} & {30.01} & {48.12} \\
{DC + FCN$^{+}$}                        & 65.51 & 44.60 & 33.61 & 50.56 \\
\hline
{OHE + FCN$^{+}$}                       & 63.27 & 44.90 & 31.87 & 48.59 \\
{OHE + DC + FCN$^{+}$}                  & \bst{66.60} & \bst{45.78} & \scd{34.28} & \bst{51.24} \\
\hline
{W2V + FCN$^{+}$}                       & {63.16} & {43.33} & {31.43} & {48.31} \\
{W2V + DC + FCN$^{+}$}                  & \scd{66.08} & \scd{45.07} & \bst{34.65} & \scd{50.96} \\
\hline
\end{tabular}
\caption{Semantic segmentation results on COCO-Stuff.}
\label{tab:experiment:cocostuff}
\vspace{-0.1in}
\end{table}


\begin{table*}[tp]
\small
\tabcolsep 2pt
\centering
\begin{tabular}{ c|ccc|ccc|ccc|ccc|ccc }
 & \multicolumn{3}{c|}{Pascal-Context} & \multicolumn{3}{c|}{ADE20K} & \multicolumn{3}{c|}{COCO-Stuff} & \multicolumn{3}{c|}{NYUDv2} & \multicolumn{3}{c}{SIFT-Flow} \\
 & pAcc & mAcc & mIU & pAcc & mAcc & mIU & pAcc & mAcc & mIU  & pAcc & mAcc & mIU  & pAcc & mAcc & mIU   \\
\hlineB{4}
{FCN$^{+}$}             & 71.25 & 53.82 & 43.19 & 77.30 & 46.94 & 36.52 
                        & 62.40 & 43.10 & 30.77 & 64.60 & 48.40 & 37.19 
                        & 87.90 & 53.04 & \bst{42.19}  \\
{DC + FCN$^{+}$}        & \bst{73.52} & \bst{56.72} & \bst{45.77} & \bst{78.03} & \bst{48.23} & \bst{37.93}
                        & \bst{65.51} & \bst{44.60} & \bst{33.61}  & \bst{65.32} & \bst{50.82} & \bst{38.76} 
                        & \bst{88.16} & \bst{55.92} & {42.11}\\
\hline
{\textit{Relative Improvement}} & 3.19\% & 5.39\%  & 5.97\%  & 0.94\%  & 2.75\%  & 3.86\% 
                        & 4.98\%  & 3.48\% & 9.23\% & 1.11\% & 5.00\% & 4.22\% 
                        & 0.30\%  & 5.42\% & -0.19\%  \\
\hline
\hline
{DilatedNet$^{+}$}      & 70.26 & 53.10 & 41.72 & 76.63 & 44.83 & 34.15 
                        & 61.06 & 41.68 & 29.62  & 62.88 & 46.13 & 34.74 
                        & {87.23} & {55.32} & {42.24} \\
{DC + DilatedNet$^{+}$} & \bst{73.47} & \bst{56.59} & \bst{45.80} & \bst{78.20} & \bst{47.89} & \bst{37.66}
                        & \bst{65.40} & \bst{45.13} & \bst{33.50} & \bst{64.14} & \bst{49.67} & \bst{37.23}
                        & \bst{88.01} & \bst{58.62} & \bst{44.14} \\
\hline
{\textit{Relative Improvement}} & 4.57\% & 6.57\% & 1.25\% & 2.05\% & 6.83\% & 10.28\% 
                        & 7.10\% & 8.28\% & 13.10\% & 2.00\% & 7.67\% & 7.17\% 
                        & 0.89\% & 3.59\% & 4.50\%  \\
\hline
\end{tabular}
\caption{Study on various semantic segmentation processes. Relative improvements are also provided to notice the differences.}
\label{tab:experiment:cross-model}
\vspace{-0.1in}
\end{table*}


We also studied another setting using the original image sizes for training and testing, where the overall performance improves with high-resolution inputs. We present these results in the appendix.

\noindent\textbf{COCO-Stuff.} We follow the standard setting of~\cite{caesar_arxiv16} in our experiments. The results are presented in Table~\ref{tab:experiment:cocostuff}. Again, it shows that DC + FCN$^{+}$ outperforms all existing methods and the FCN$^{+}$ baseline considerably.

\noindent\textbf{Discussion.} The above experiments show that DC + FCN$^{+}$ can achieve state-of-the-art performance. SPP + FCN$^{+}$ is slightly worse, but still achieves reasonable results (especially on PASCAL-Context). DC works better than SPP because DC is an ensemble-like method that combines location-aware predictions. For the rest of the experiments, we use DC as the default architecture for inferring LabelBank from visual appearance.


\vspace{-0.1in}
\subsubsection{Inference from Meta Data}
\label{sec:experiment:labelbank:meta}
\vspace{-0.05in}

Next we conduct experiments to infer LabelBank from meta data, where we leverage image attributes and textual descriptions to help discovering holistic concepts in images.


Specifically, we extract image attributes on ADE20K and COCO-Stuff. We rely on the provided taxonomies of the semantic object categories in the two datasets. For each object category of an image, we assign attributes to be its ancestor hypernyms. The image attributes are collected as the union of attributes for all semantic categories present in the image. We then apply the OHE architecture to infer the LabelBank representation.

Textual descriptions are available on COCO-Stuff, and we apply our W2V architecture for LabelBank inference. In detail, we first perform GloVe embedding~\cite{pennington_emnlp14} on individual words, and then obtain a feature vector of the textual description by averaging the embeddings of all words. Advanced embedding techniques (\textit{e.g.}, skip-thought vector~\cite{kiros_nips15} or paragraph vector~\cite{le_icml14}) are applicable, and we leave these for future exploration.

The results are provided in Tables~\ref{tab:experiment:ade-384} and \ref{tab:experiment:cocostuff}. Our methods, OHE + FCN$^{+}$ and W2V + FCN$^{+}$, provide a clear improvement over the baseline FCN$^{+}$. It again verifies the generalizability of our framework for LabelBank inference.

\vspace{-0.1in}
\subsubsection{Combining Meta Data and Visual Appearance}
\label{sec:experiment:labelbank:metavisual}
\vspace{-0.05in}

Furthermore, we could use meta data and visual appearance together for a more accurate inference of the LabelBank representation. In our implementation, we simply append the OHE/W2V embedded features to each location of the feature map of our DC architecture. This results in the OHE + DC + FCN$^{+}$ and W2V + DC + FCN$^{+}$ methods, and we report their performance in Tables~\ref{tab:experiment:ade-384} and \ref{tab:experiment:cocostuff}.

It clearly shows that OHE + DC + FCN$^{+}$ and W2V + DC + FCN$^{+}$ achieve the best performance on ADE20K and COCO-Stuff. Also note that combining meta data and visual appearance together generates better LabelBank than using individual data sources -- OHE + DC + FCN$^{+}$ outperforms OHE + FCN$^{+}$ and DC + FCN$^{+}$ on the two datasets.

\subsection{Study on the Semantic Segmentation Process}
\label{sec:experiment:segmentation}
\vspace{-0.05in}

Here we verify the generalizability of our framework over different semantic segmentation processes. We use DC for LabelBank inference, and instantiate the segmentation process with FCN$^{+}$ and DilatedNet$^{+}$.


The comparative results are shown in Table~\ref{tab:experiment:cross-model}. Please refer to Tables~\ref{tab:experiment:pascal}, \ref{tab:experiment:ade-384} and \ref{tab:experiment:cocostuff} for a complete comparison with existing approaches in literature. As shown, DC + FCN$^{+}$ and DC + DilatedNet$^{+}$ consistently improve over their baseline counterparts FCN$^{+}$ and DilatedNet$^{+}$ (except on SIFT-Flow, DC + FCN$^{+}$ performs slightly worse than FCN$^{+}$). This validates that our framework is general and can improve state-of-the-art CNN-based semantic segmentation networks.



\vspace{-0.1in}
\section{Analysis}
\label{sec:analysis}
\vspace{-0.05in}

In this section, we conduct empirical studies to further understand our framework. We asked the following two questions: (i) what is the maximum performance gain we could achieve if we have perfect LabelBank inference? and (ii) how could we refine the current LabelBank inference to further improve the performance? We provide answers and insights on these two questions in the following.


\vspace{-0.05in}
\subsection{Oracle LabelBank}
\label{sec:analysis:oracle}
\vspace{-0.05in}

Our experiments in Section~\ref{sec:experiment} show that semantic segmentation benefits from the guidance of the LabelBank. However, by no means would we claim that our LabelBank inference is perfect. Therefore, an interesting question to raise is: how well could our framework perform if we had perfect LabelBank inference?

We answer this question by replacing our LabelBank inference process with a static process that returns ``oracle'' LabelBank. The oracle LabelBank can be simply derived from the ground-truth pixel labels, by taking the value of infinity if a semantic class is present in the image, and negative infinity otherwise. We keep the rest of the framework intact, and train it end-to-end towards optimizing the segmentation loss (see Section~\ref{sec:framework:inference:train} for details).

\begin{figure}[tp]
\vspace{-0.1in}
\begin{center}
\includegraphics[width=0.475\textwidth]{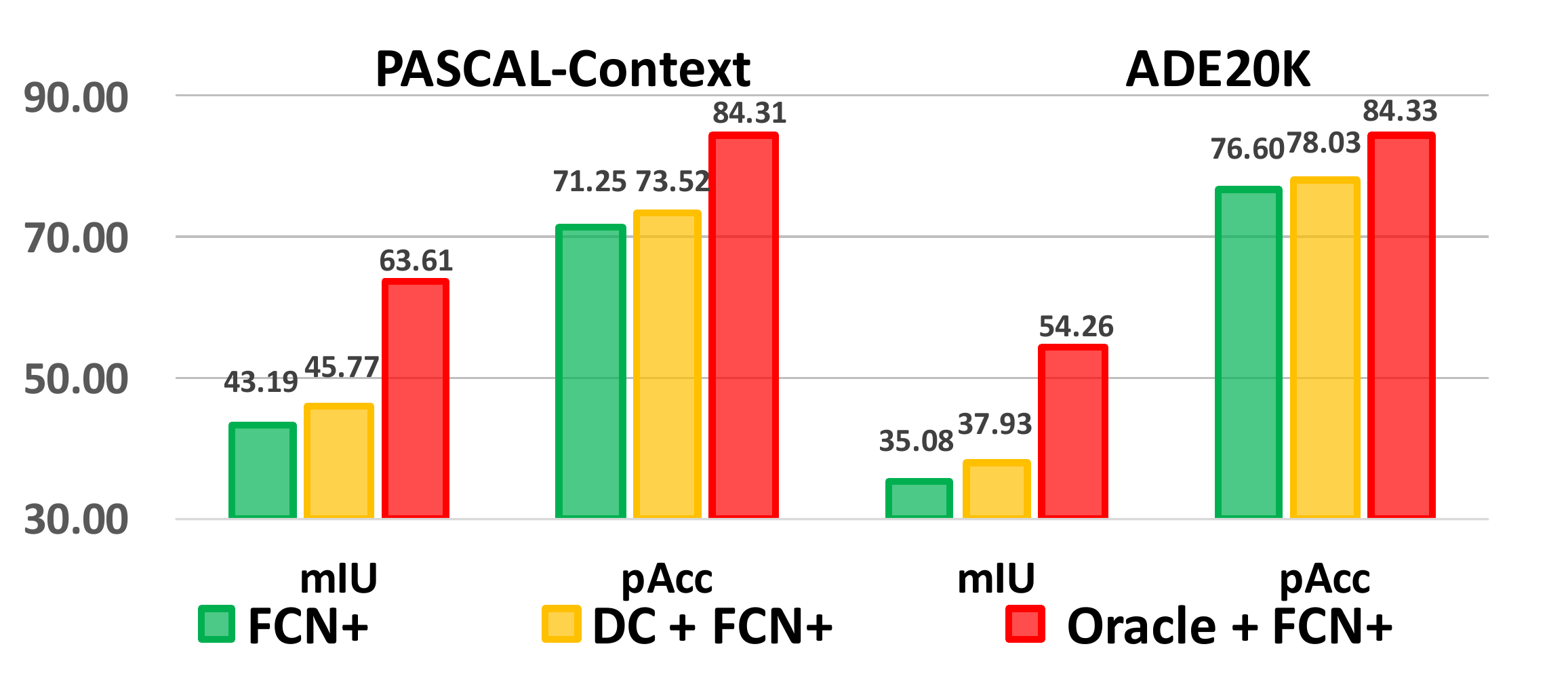}
\end{center}
\caption{The pAcc and mIU performance of FCN$^{+}$, DC + FCN$^{+}$ and Oracle + FCN$^{+}$ on PASCAL-Context and ADE20K. }
\label{fig:analysis:oracle}
\vspace{-0.05in}
\end{figure}

We compare FCN$^{+}$, DC + FCN$^{+}$, and Oracle + FCN$^{+}$ in Figure~\ref{fig:analysis:oracle}. It clearly shows that Oracle + FCN$^{+}$ using perfect LabelBank boosts the semantic segmentation performance significantly -- for example, the mIU values are improved to 63.61\% on PASCAL-Context and 54.26\% on ADE20K. We believe that Oracle + FCN$^{+}$ provides the upper-bound on performance we could possibly achieve with an FCN$^{+}$ segmentation network with our framework. We would gradually approach this upper-bound as we obtain a better and better LabelBank inference process.

\subsection{Noisy LabelBank}
\label{sec:analysis:noisy}
\vspace{-0.05in}

Certainly we could never assume oracle LabelBank in real scenarios. For example, when applying a threshold of 0 on the LabelBank confidences derived by DC + FCN$^{+}$, we observe 90.93\% recall and 46.75\% precision on PASCAL-Context images, and 69.07\% recall and 47.59\% precision on ADE20K images. We believe that precision and recall are two key measures on the goodness of the LabelBank representation. So here we study the impact of LabelBank precision and recall on the semantic segmentation performance, and provide insight on how to refine the current inference process for further performance gain.

We have evaluated our framework with various precision and recall settings on the LabelBank. We start with the oracle LabelBank, and contaminate it to certain precision and recall levels. Specifically, we degrade the precision by adding in more and more noisy image labels, and degrade the recall by removing more and more ground-truth image labels. We experimented with a FCN$^{+}$ based semantic segmentation network, and leveraged the noisy LabelBank in holistic filtering. Due to computational resource limitations, we did not retrain the FCN$^{+}$ networks (as we have done for Oracle + FCN$^{+}$). Detailed experimental setup is provided in the supplementary material.

\begin{figure}[tp]
\vspace{-0.05in}
\begin{center}
\includegraphics[width=0.425\textwidth]{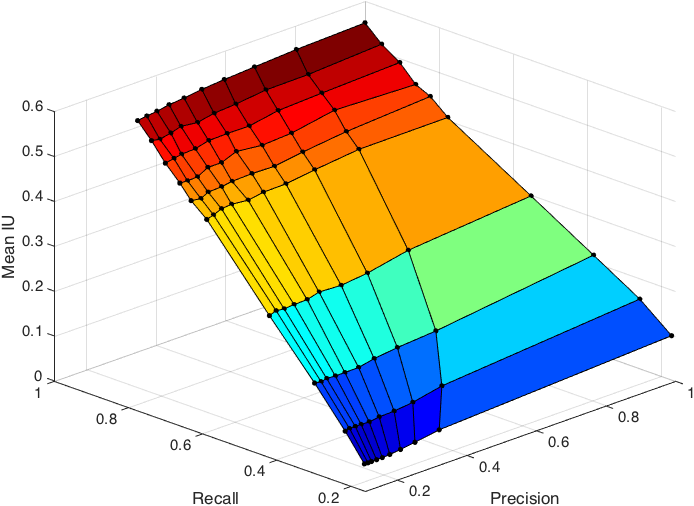}
\end{center}
\caption{The mean IU grids of our framework with respect to various precisions and recalls, evaluated on PASCAL-Context and using FCN$^{+}$ based semantic segmentation network.}
\label{fig:analysis:grid}
\vspace{-0.05in}
\end{figure}

Figure~\ref{fig:analysis:grid} plots the mean IU grid. It is shown that recall matters much more than precision -- the mean IU decreases more significantly with respect to the degradation of recall than precision. Note that this observation is obtained with noisy binary labels in the LabelBank, and may not apply directly to continuous LabelBank confidences. However, it does suggest a promising direction for future development in LabelBank inference -- it could be beneficial to push hard on recalling ground-truth image labels.

\vspace{-0.1in}
\section{Conclusion}
\label{sec:conclusion}
\vspace{-0.05in}

This paper motivates the use of a holistic LabelBank representation for semantic segmentation. We have presented a generic framework consisting of three components: LabelBank inference, semantic segmentation, and holistic filtering. The LabelBank inference process derives a holistic LabelBank representation of an image from various data sources and inference architectures. The semantic segmentation applies state-of-the-art CNN-based networks to generate a preliminary segmentation map. Finally, the holistic filtering process refines segmentation results by leveraging the LabelBank information to filter out false-positive pixel predictions. Experiments on benchmark semantic segmentation datasets show the effectiveness of the proposed framework. We believe that our solution is general and could be applied to many other applications, for example, improving object detection with holistic LabelBank, \textit{etc}.


{\small
\bibliographystyle{ieee}
\bibliography{ref}
}

\begin{appendices}

We provide details to support our main paper in this appendix. The organization is as follows. Section~\ref{sec:sup:arch} gives details on the implementation of our framework. Section~\ref{sec:sup:train} specifies our training strategy. Section~\ref{sec:sup:dataset} describes the experimental datasets. Section~\ref{sec:sup:result} reports additional experimental results. Section~\ref{sec:sup:feature} studies the feature network. Section~\ref{sec:sup:noisy} explains the precision and recall settings for obtaining noisy LabelBank. Finally, Section~\ref{sec:sup:viz} provides additional qualitative visualizations.

\section{Implementation Details}
\label{sec:sup:arch}

In this section, we explain the detailed implementation of FCN$^+$ and DilatedNet$^+$ in Section~\ref{sec:sup:arch:plus}, the two-layer CNN used in the DC architecture in Section~\ref{sec:sup:arch:dc}, multi-layer perceptron (MLP) in Section~\ref{sec:sup:arch:mlp}, and our feature network in Section~\ref{sec:sup:arch:feature}.

\subsection{FCN$^+$ and DilatedNet$^+$}
\label{sec:sup:arch:plus}

\begin{figure*}[htp]
\centering
\includegraphics[width=1\textwidth]{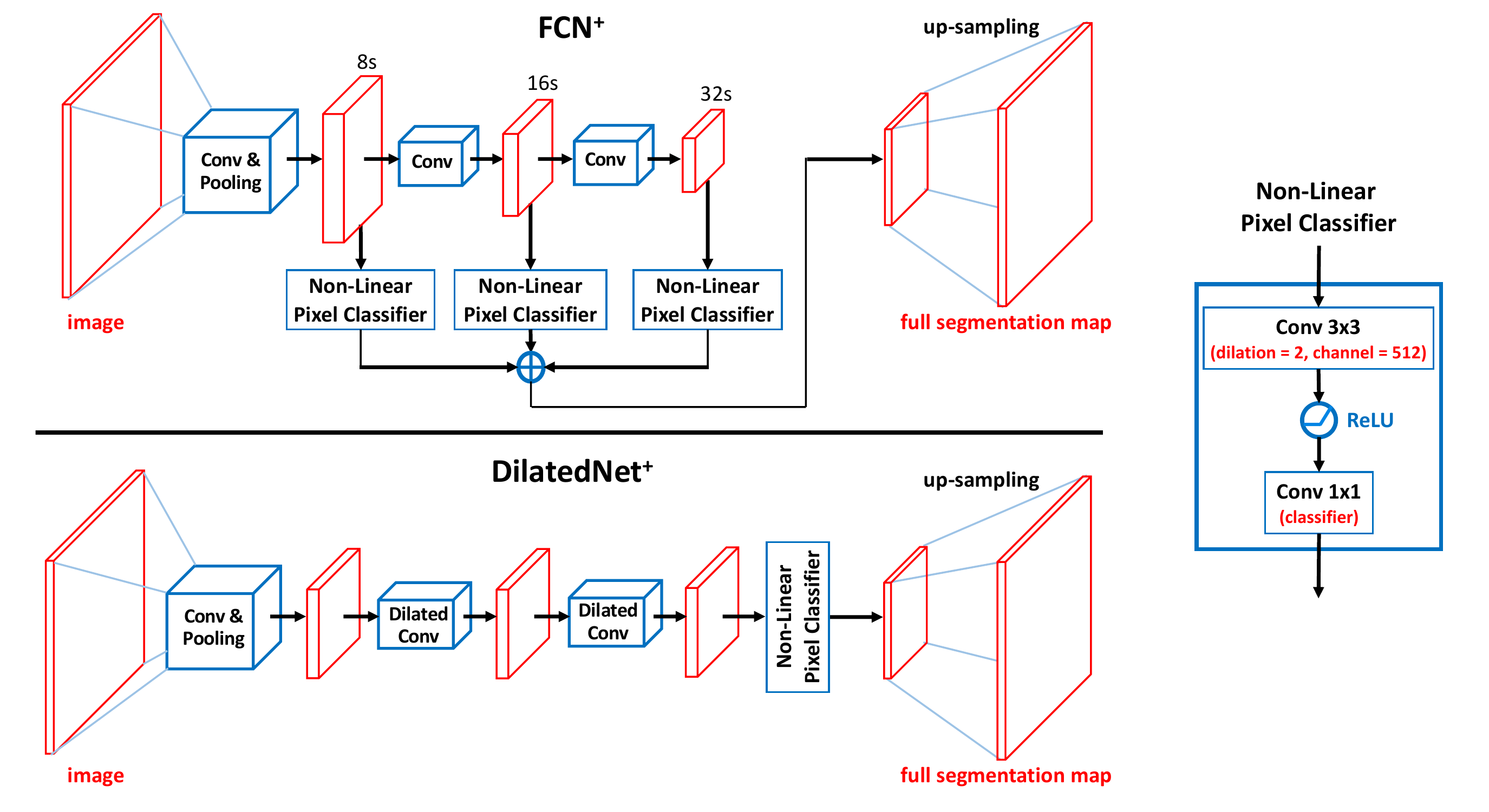}
\caption{The network architectures of FCN$^{+}$ and DilatedNet$^{+}$. Note that the only modification from FCN~\cite{shelhamer_pami16} and DilatedNet~\cite{chen_arxiv16} is to use our designed non-linear pixel classifier to generate segmentation maps. More details are described in the text.}
\label{fig:sup:arch:plus}
\end{figure*}

The semantic segmentation process is responsible for generating a preliminary segmentation map for the holistic filtering process to refine under the guidance of LabelBank. As mentioned in Section~3, we have slightly modified FCN~\cite{shelhamer_pami16} and DilatedNet~\cite{chen_arxiv16} to serve as our semantic segmentation process, \textit{i.e.}, FCN$^{+}$ and DilatedNet$^{+}$. We show the detailed network architectures in Figure~\ref{fig:sup:arch:plus}, and describe the details in the following.

FCN$^{+}$ has a similar structure as FCN~\cite{shelhamer_pami16}: it first applies a feature network to obtain a feature map on an image, then uses skip features to generate three segmentation maps of various down-sampling rates (8s, 16s and 32s), then fuses them together, and finally up-samples to the original image size for semantic segmentation. The only modification is that FCN$^{+}$ applies our designed \textit{non-linear pixel classifier} (instead of the linear pixel classifier in FCN) to generate a segmentation map from the input feature map.

The non-linear pixel classifier leverages non-linearity and dilated convolution to model pixel labeling. It consists of a dilated convolutional layer~\cite{yu_iclr16,chen_arxiv16} to aggregate contextual information, a ReLU layer for non-linearity, as well as a one-by-one convolutional layer to predict the segmentation map. In our experiments, we apply 3 by 3 convolutional kernels in the dilated convolutional layer, using a dilation rate of 2 under 512 channels.

Our DilatedNet$^{+}$ is derived from DilatedNet~\cite{chen_arxiv16} with the same modification -- we apply our non-linear pixel classifier on the feature map (obtained after dilated convolutions) to generate the segmentation map. We kept the rest of the structure the same as~\cite{chen_arxiv16}.

\begin{table}[tp]
\small
\tabcolsep 8pt
\centering
\begin{tabular}{ c|cccc }
 & pAcc & mAcc & mIU & fwIU \\ \hlineB{4} 
{FCN~\cite{shelhamer_pami16}} & 67.5 & \scd{52.3} & 39.1 & 53.0 \\
{FCN$^{+}$} & \bst{68.64} & \bst{52.36} & \bst{39.45} & \bst{54.50} \\
\hline
{DilatedNet~\cite{chen_arxiv16}} & - & - & 37.6 & - \\
{DilatedNet$^{+}$} & \scd{68.28} & \scd{52.30} & \scd{39.21} & \scd{53.75} \\
\hline
\end{tabular}
\caption{Comparison of our modified semantic segmentation networks with the original FCN~\cite{shelhamer_pami16} and DilatedNet~\cite{chen_arxiv16}. The results are reported on the PASCAL-Context dataset, using the same feature network derived from the 16-layer VGGNet~\cite{simonyan_iclr15}.}
\label{tab:sup:plus}
\end{table}

We show an empirical comparison between the original FCN/DilatedNet and our FCN$^{+}$/DilatedNet$^{+}$ in Table~\ref{tab:sup:plus}. For a fair comparison, we apply the same feature network, \textit{i.e.}, 16-layer VGGNet~\cite{simonyan_iclr15}, for all methods. Our networks obtained slightly better performance than the original methods due to the usage of the non-linear pixel classifier. 


\subsection{Two-Layer CNN in the DC Architecture}
\label{sec:sup:arch:dc}

As mentioned in Section~3.2.2 and Figure~3, we use a two-layer CNN to infer the LabelBank representation from densely cropped image windows. The two-layer CNN is structured as follows. The first is a dilated convolutional layer. We use $k \times k$ convolutional kernels with a dilation rate of $r$ under $d$ channels. In our experiments, we empirically set the patch size as 224, $k=3$, $r=2$ and $d=512$. We apply ReLU activation on its outputs for non-linearity. The second layer is a one-by-one convolutional layer to predict the current window's LabelBank representation. We then max-pool the location-aware predictions to compose the final LabelBank representation for the image.

\subsection{Multi-Layer Perceptron (MLP)}
\label{sec:sup:arch:mlp}

We apply the same MLP structure in our SPP, OHE and W2V architectures for LabelBank inference. It first employs a fully-connected layer to map the input vector to (2048) hidden units, then applies a ReLU activation function, and finally leverages another fully-connected layer to generate the output representation, \textit{i.e.}, the $k$-dimensional LabelBank.

\subsection{Feature Network}
\label{sec:sup:arch:feature}

When inferring the LabelBank from visual appearance, our SPP and DC architectures both employ a feature network (\textit{i.e.}, low-level layers of convolution and pooling) to extract a feature map on a given image. In our experiments, we empirically build the feature network from the 152-layer ResNet~\cite{he_cvpr16} by removing the top fully-connected layers and keeping the convolutional and pooling layers.

Also note that there is a feature network in our semantic segmentation process (\textit{i.e.}, FCN$^{+}$ and DilatedNet$^{+}$). In our experiments, we share the same feature network across LabelBank inference and semantic segmentation for computational efficiency and feature generalizability.

With the shared feature network, our framework looks similar to multi-task learning of classification and segmentation. However, we point out that the key to success lies in the LabelBank based holistic filtering. We provide empirical evidence to support this in Section~\ref{sec:sup:feature:multitask}. Furthermore, we also conduct experiments in Section~\ref{sec:sup:feature:variant} to show that our framework is flexible and can take various feature networks.

\section{Training Strategy}
\label{sec:sup:train}

\noindent\textbf{Optimization.} We follow the practice of FCN~\cite{shelhamer_pami16} to train our framework -- optimizing the objective by stochastic gradient descent with a small batch size (\textit{e.g.}, 1) and a large momentum (\textit{e.g.}, 0.99). We train for approximately 60 epochs and choose the best models through validation.

\noindent\textbf{Data augmentation.} It has been shown that data augmentation is a practical technique to boost semantic segmentation performance. In our experiments, we have applied horizontal flipping as well as scale augmentation when training our networks. For scale augmentation, we randomly pick a scale factor in the set $\{0.7, 0.8, 0.9, 1.0, 1.1, 1.2, 1.3\}$ to scale each image.

\section{Experimental Datasets}
\label{sec:sup:dataset}

\noindent\textbf{PASCAL-Context} is an extension of the PASCAL VOC 2010 dataset, with detailed pixel-wise annotations~\cite{mottaghi_cvpr14}. The semantic labels include both objects and stuff present in the image. Following~\cite{mottaghi_cvpr14,shelhamer_pami16}, we evaluate our network on the most frequent 59 classes alongside one background class. The training and testing sets contain 4,998 and 5,105 images, respectively.

\noindent\textbf{ADE20K} is another dataset with densely annotated objects and stuff. We learn our model on the 20,210 training images, and report performance on the 2,000 validation images. We did not evaluate on its 5,000 test images as the ground-truth annotations are not publicly available. Following~\cite{zhou_arxiv16}, we select the top 150 semantic classes ranked by their total pixel ratios, including 35 object classes and 115 stuff classes. The pixels from the 150 classes occupy 92.75\% of all pixels in the dataset.

\noindent\textbf{COCO-Stuff} is a recent densely annotated dataset~\cite{caesar_arxiv16} with images sampled from COCO~\cite{lin_eccv14}. There are 80 thing and 91 stuff classes. Following the standard train/test split, we train on 9,000 images and test on 1,000 images.

\noindent\textbf{NYUDv2} is an RGB-D dataset on indoor scenes collected using Microsoft Kinect. It has 1,449 RGB-D images, with pixel-wise labels that have been coalesced into 40 semantic classes by Gupta \textit{et al.}~\cite{gupta_cvpr13}. We experiment with the standard split of 795 training images and 654 testing images.

\noindent\textbf{SIFT-Flow} dataset contains 2,688 images thoroughly annotated by LabelMe users~\cite{liu_pami11,liu_pami11a} with 33 semantic pixel labels (\textit{e.g.}, mountain, sun, bridge, \textit{etc}). We use the same split as~\cite{liu_pami11,liu_pami11a} -- 2,488 images for training and 200 images for testing.

\section{Additional Results}
\label{sec:sup:result}

Due to the space limit in our main paper, we defer our experimental results on SIFT-Flow and NYUDv2 here. These two datasets have no additional meta data beyond the image visual appearance. So here we compare our SPP + FCN$^{+}$ and DC + FCN$^{+}$ with the baseline FCN$^{+}$ as well as state-of-the-art approaches. The details are described as follows.

\subsection{SIFT-Flow}
\label{sec:sup:result:siftflow}

The results are reported in Table~\ref{tab:sup:result:siftflow}, which shows that our methods perform the best over all the compared methods. Note that the current state-of-the-art FCN-8s~\cite{shelhamer_pami16} leverages the available pixel-wise geometric labels (\textit{i.e.}, horizontal, vertical and sky) as extra supervision, whereas our methods do not use them. These observations show the utility of our framework in semantic segmentation.

\begin{table}[tp]
\small
\tabcolsep 3pt
\centering
\begin{tabular}{ c|cccc }
 & pAcc & mAcc & mIU & fwIU \\
\hlineB{4}
{Liu~\textit{et al.}~\cite{liu_pami11a}}    & 76.7        & -           & -                 & -      \\
{Exemplar SVM~\cite{tighe_cvpr13}}          & 75.6        & 41.1        & -                 & -      \\
{SVM + MRF~\cite{tighe_cvpr13}}             & 78.6        & 39.2        & -                 & -      \\
{Multiscale CNN + Natural~\cite{farabet_pami13}} & 72.3   & 50.8        & -                 & -      \\
{Multiscale CNN + Balanced~\cite{farabet_pami13}} & 78.5  & 29.6        & -                 & -      \\
{Recurrent CNN~\cite{pinheiro_icml14}}      & 77.7        & 29.8        & -                 & -      \\
{ParseNet~\cite{liu_iclr16}}                & 86.8        & 52.0        & 40.4              & 78.1   \\
{FCN-8s~\cite{shelhamer_pami16}}            & 85.9        & 53.9        & 41.2              & 77.2   \\
\hline
{FCN$^{+}$}                                 & {87.90} & {53.04}  & \scd{42.19}  & {79.80}         \\
{SPP + FCN$^{+}$}                           & \bst{88.20} & \scd{55.91} & \bst{45.00} & \scd{80.22} \\
{DC + FCN$^{+}$}                            & \scd{88.16} & \bst{55.92} & {42.11} & \bst{80.51} \\

\hline
\end{tabular}
\caption{Semantic segmentation results on SIFT-Flow.}
\label{tab:sup:result:siftflow}
\end{table}

It is worth mentioning that a competitive method on the SIFT-Flow dataset is proposed by Lin~\textit{et al.}~\cite{lin_cvpr16}, achieving 88.1\% pAcc, 53.4\% mAcc and 44.9\% mIU. However, this method up-samples training and testing images by a factor of two, and benefits from the high-resolution images to obtain superior performance. On the other hand, all the other methods including ours do not up-sample images, and thus are not directly comparable with~\cite{lin_cvpr16}.

\subsection{NYUDv2}
\label{sec:sup:result:nyud}

We provide the evaluation results in Table~\ref{tab:sup:result:nyud}. It is shown that our methods achieve the best performance, especially with DC + FCN$^{+}$.

\begin{table}[tp]
\small
\tabcolsep 4pt
\centering
\begin{tabular}{ c|cccc }
 & pAcc & mAcc & mIU & fwIU \\
\hlineB{4}
{Gupta~\textit{et al.}~\cite{gupta_eccv14}}         & 60.3          & -         & 28.6      & 47.0      \\
{FCN-32s + RGB~\cite{shelhamer_pami16}}             & 61.8          & 44.7      & 31.6      & 46.0      \\ 
{FCN-32s + RGB + D~\cite{shelhamer_pami16}}         & 62.1          & 44.8      & 31.7      & 46.3      \\ 
{FCN-32s + RGB + HHA~\cite{shelhamer_pami16}}       & \scd{65.3}    & 44.0      & 33.3      & 48.6      \\ 
\hline
{FCN$^{+}$} & {64.60} & {48.40} & {37.19} & {49.41}  \\ 
{SPP + FCN$^{+}$} & 65.05 & \bst{51.99} & \bst{38.79} & \scd{50.11} \\
{DC + FCN$^{+}$}  & \bst{65.32} & \scd{50.82} & \scd{38.76} & \bst{50.25} \\ 
\hline
\end{tabular}
\caption{Semantic segmentation results on NYUDv2.}
\label{tab:sup:result:nyud}
\end{table}

Our methods only use the color images for training, ignoring the depth information. DC + FCN$^{+}$ still improves 5.46\% mIU over the state-of-the-art method, FCN-32s + RGB + HHA~\cite{shelhamer_pami16}, which utilizes both the color and depth information. It again verifies the effectiveness of the proposed framework.

\section{Ablation Studies on Feature Network}
\label{sec:sup:feature}

In this section, we conduct ablation studies on our feature network. First, in Section~\ref{sec:sup:feature:multitask}, we compare our framework with multi-task learning that also uses a shared feature network. Second, in Section~\ref{sec:sup:feature:variant}, we evaluate the flexibility of our framework against various feature networks.

\subsection{Comparison with Multi-Task Learning}
\label{sec:sup:feature:multitask}

\begin{table}[tp]
\small
\tabcolsep 8pt
\centering
\begin{tabular}{ c|cccc }
Method & pAcc & mAcc & mIU & fwIU \\
\hlineB{4}
{FCN$^{+}$}                           & 71.25       & 53.82       & 43.19           & 57.63 \\ 
{Multi-Task FCN$^{+}$}                & \scd{71.43} & \scd{54.15} & \scd{43.40} & \scd{57.89} \\  
{DC + FCN$^{+}$}                      & \bst{73.52} & \bst{56.72} & \bst{45.77} & \bst{60.05} \\
\hline
\end{tabular}
\caption{Comparison with multi-task learning on PASCAL-Context.}
\label{tab:sup:feature:multitask}
\end{table}

\begin{table*}[tp]
\small
\tabcolsep 15pt
\centering
\begin{tabular}{ c|c|cccc }
 & Feature Network & pAcc & mAcc & mIU & fwIU \\ \hlineB{4} 
{FCN$^{+}$}                 & 16-layer VGGNet     & 68.64 & 52.36 & 39.45 & 54.50 \\ 
{DC + FCN$^{+}$}        & 16-layer VGGNet     & 69.24 & 51.81 & 40.11 & 55.14 \\
{DilatedNet$^{+}$}          & 16-layer VGGNet     & 68.28 & 52.30 & 39.21 & 53.75 \\ 
{DC + DilatedNet$^{+}$} & 16-layer VGGNet     & 70.50 & 54.24 & 41.83 & 56.37 \\ \hlineB{4}
{FCN$^{+}$}                 & 152-layer ResNet    & 71.25 & 53.82 & 43.19 & 57.63 \\
{DC + FCN$^{+}$}        & 152-layer ResNet    & \bst{73.52} & \bst{56.72} & \scd{45.77} & \scd{60.05} \\
{DilatedNet$^{+}$}          & 152-layer ResNet    & 70.26 & 53.10 & 41.72 & 56.57 \\ 
{DC + DilatedNet$^{+}$} & 152-layer ResNet    & \scd{73.47} & \scd{56.59} & \bst{45.80} & \bst{60.23} \\
\hline
\end{tabular}
\caption{Our framework using various feature networks on the PASCAL-Context dataset. We highlight the best performance in red and boldfaced, and the $2^{nd}$ best in blue and underline.}
\label{tab:sup:feature:pascal}
\end{table*}

In our framework implementation, we share the feature network across the visual appearance based LabelBank inference and the semantic segmentation process. The resultant framework has a similar structure as multi-task learning of classification and segmentation. However, we emphasize that our performance gain mainly comes from the LabelBank based holistic filtering, not the generic features learned from the shared feature network. We conduct experiments to support this in the following.

We have evaluated a standard multi-task learning method, where we have a classification branch in parallel with a semantic segmentation branch, with a shared feature network but without LabelBank based holistic filtering. We instantiate the classification and segmentation branches with DC and FCN$^{+}$, respectively. The results on PASCAL-Context are reported in Table~\ref{tab:sup:feature:multitask}, which shows no significant performance gain with the multi-task training over the baseline FCN$^{+}$. In contrast, our DC + FCN$^{+}$ outperforms Multi-Task FCN$^{+}$ and FCN$^{+}$ substantially. It validates the critical usage of our LabelBank based holistic filtering for semantic segmentation.

It is also interesting to note that Multi-Task FCN$^{+}$ has exactly the same network capacity (\textit{i.e.}, number of parameters) as our DC + FCN$^{+}$, but it performs worse than ours. It shows that increasing network capacity might not necessarily increase performance. Instead, it is our LabelBank based holistic filtering that boosts semantic segmentation.


\subsection{Feature Network Variants}
\label{sec:sup:feature:variant}

We have also conducted experiments to study the flexibility of our framework in taking various feature networks. In detail, we have tried the 16-layer VGGNet~\cite{simonyan_iclr15} and the 152-layer ResNet~\cite{he_cvpr16} respectively as our feature network. We evaluate our methods (\textit{i.e.}, DC + FCN$^{+}$ and DC + DilatedNet$^{+}$) and our baselines (\textit{i.e.}, FCN$^{+}$ and DilatedNet$^{+}$) accordingly. The comparative results on the PASCAL-Context dataset are reported in Table~\ref{tab:sup:feature:pascal}.

The table clearly shows that our methods improve over the baselines, using either VGGNet or ResNet. It verifies that our framework is flexible and can adopt various feature networks to improve semantic segmentation. This flexibility enables our framework to take advantage of the continual improvement in CNN architectures. Furthermore, it is beneficial to use ResNet as our feature network as it consistently outperforms its VGGNet counterpart. This is reasonable since ResNet employs a deeper structure than VGGNet to capture rich image features.

\section{Settings for Noisy LabelBank Experiments}
\label{sec:sup:noisy}

In Section~5.2, we experimented with various precision and recall settings on LabelBank. The detailed setup is as follows.

To degrade the precision, we have added in $n_{p} \in \{1, 2, \cdots, 10\}$ noisy labels per image. Similarly, to degrade the recall, we have removed $n_{r} \in \{1, 2, \cdots, 10\}$ ground-truth labels per image (if there are fewer than $n_{r}$ ground-truth labels in an image, we just remove them all). Note that $n_{p}$ and $n_{r}$ can be set as fractional values to further refine the numerical accuracy of precision and recall. For example, setting $n_{r}=2.3$ means that we will first remove 2 (the integer part of $n_{r}$) ground-truth labels on each image, and then randomly pick up 30\% (the decimal part of $n_{r}$) of the images to remove one more ground-truth label each. With this trick, we also tried $n_{r} \in \{0.2, 0.4, 0.6, 0.8\}$ in our experiment. We conduct holistic filtering for semantic segmentation with an exhaustive grid search of $n_{p}$ and $n_{r}$ values. Note that each combination of $n_{p}$ and $n_{r}$ produces LabelBank with certain precision and recall of ground-truth labels (by averaging over all images). The mean IU grid is plotted in Figure~6.

\section{Visualizations}
\label{sec:sup:viz}

We select sample images from PASCAL-Context and ADE20K, and visualize the semantic segmentation results in Figures~\ref{fig:sup:viz:pascal} and \ref{fig:sup:viz:ade}. Here we use the DC architecture to infer LabelBank from image visual appearance. As a comparison with existing methods, we have also provided the FCN~\cite{shelhamer_pami16} results on PASCAL-Context and the DilatedNet~\cite{zhou_arxiv16} results on ADE20K. The qualitative comparison verifies the utility of LabelBank for semantic segmentation -- it helps to filter out false-positive pixel predictions via the holistic filtering process.

\def\VizImageWidth{0.835in}
\def\VizImageHeightLong{1.2in}
\def\VizImageHeightMiddle{0.835in}
\def\VizImageHeightShort{0.7in}

\begin{figure*}[tp]
\small
\centering
\tabcolsep 1pt
\begin{tabular}{cccc|cccc}
Image & FCN~\cite{shelhamer_pami16} & DC + FCN$^{+}$ & Ground Truth & Image & FCN~\cite{shelhamer_pami16} & DC + FCN$^{+}$ & Ground Truth \\
\includegraphics[width=\VizImageWidth,height=\VizImageHeightLong]{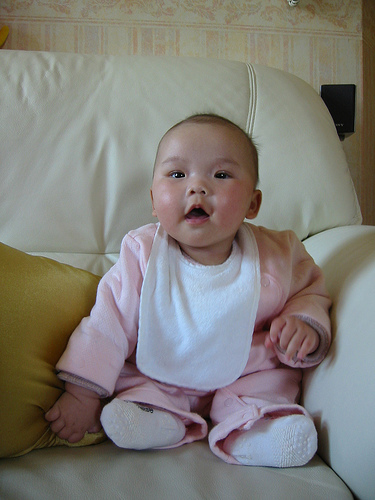} &
\includegraphics[width=\VizImageWidth,height=\VizImageHeightLong]{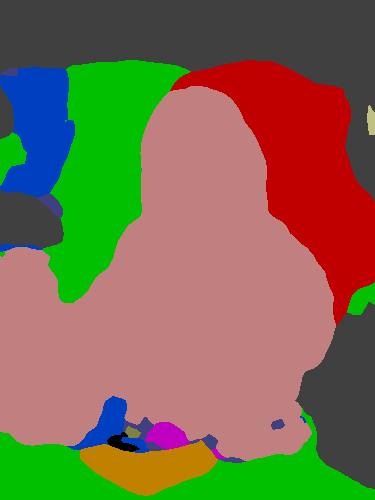} &
\includegraphics[width=\VizImageWidth,height=\VizImageHeightLong]{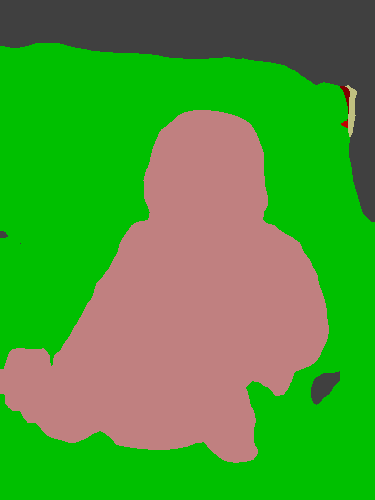} &
\includegraphics[width=\VizImageWidth,height=\VizImageHeightLong]{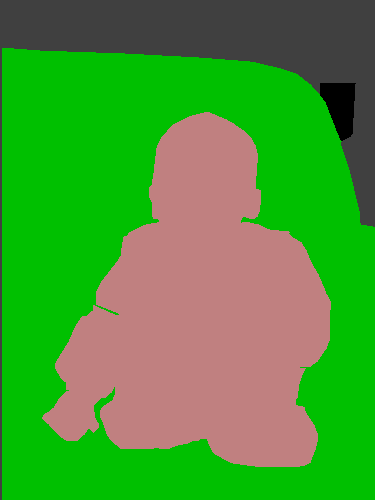} &
\includegraphics[width=\VizImageWidth,height=\VizImageHeightLong]{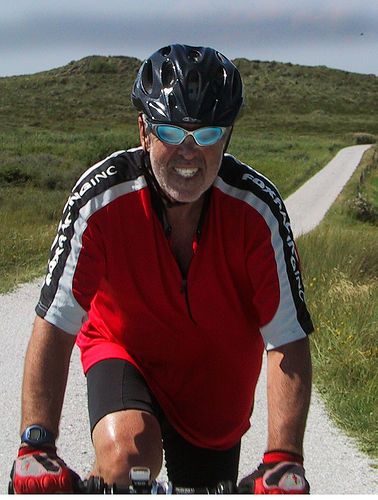} &
\includegraphics[width=\VizImageWidth,height=\VizImageHeightLong]{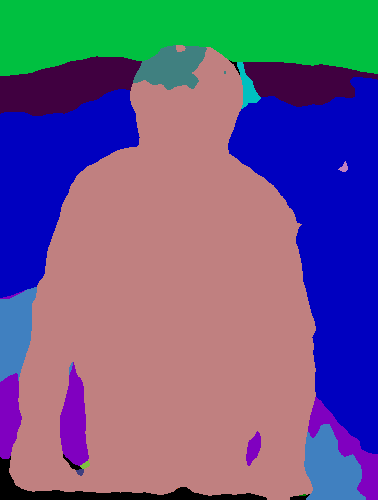} &
\includegraphics[width=\VizImageWidth,height=\VizImageHeightLong]{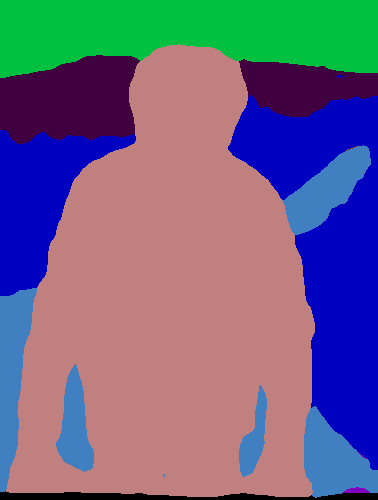} &
\includegraphics[width=\VizImageWidth,height=\VizImageHeightLong]{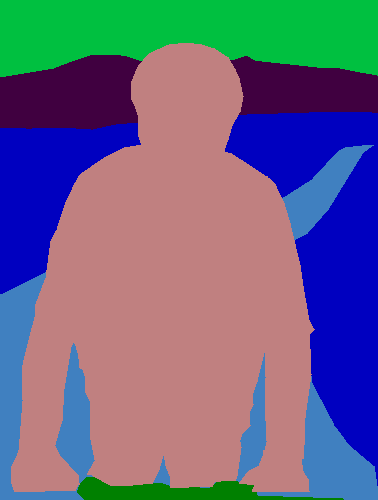} \\
\includegraphics[width=\VizImageWidth,height=\VizImageHeightLong]{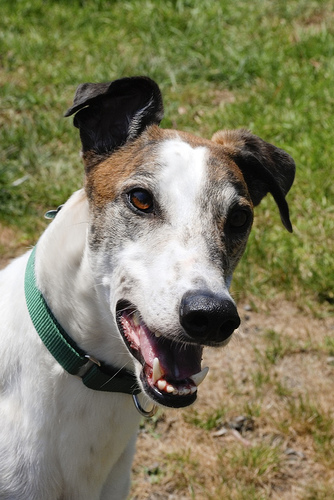} &
\includegraphics[width=\VizImageWidth,height=\VizImageHeightLong]{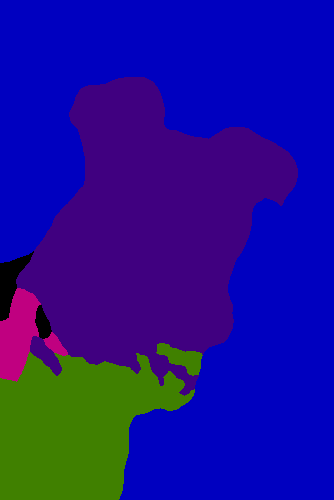} &
\includegraphics[width=\VizImageWidth,height=\VizImageHeightLong]{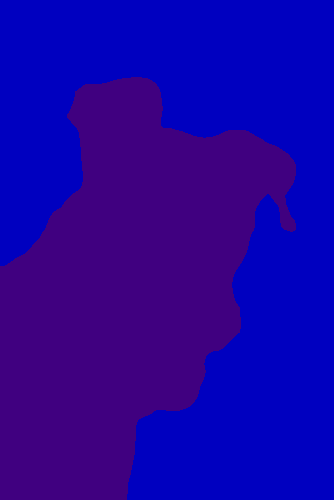} &
\includegraphics[width=\VizImageWidth,height=\VizImageHeightLong]{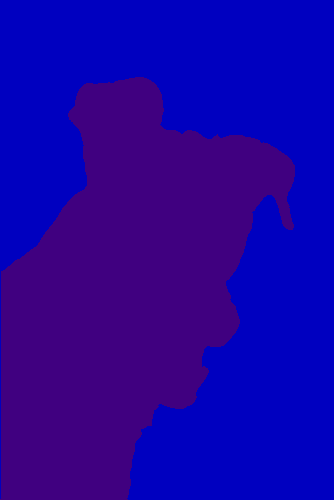} &
\includegraphics[width=\VizImageWidth,height=\VizImageHeightLong]{file/pascal/2008_002383.jpg} &
\includegraphics[width=\VizImageWidth,height=\VizImageHeightLong]{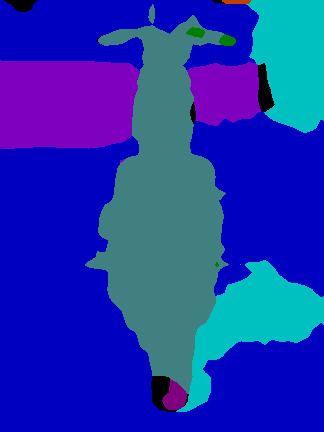} &
\includegraphics[width=\VizImageWidth,height=\VizImageHeightLong]{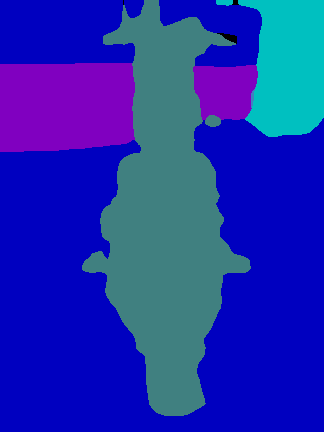} &
\includegraphics[width=\VizImageWidth,height=\VizImageHeightLong]{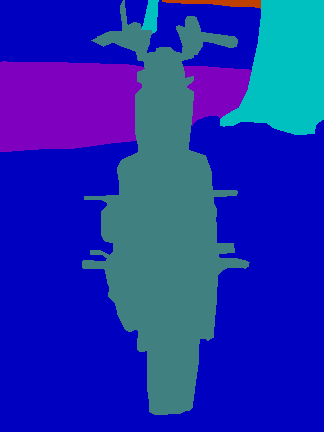} \\
\includegraphics[width=\VizImageWidth,height=\VizImageHeightShort]{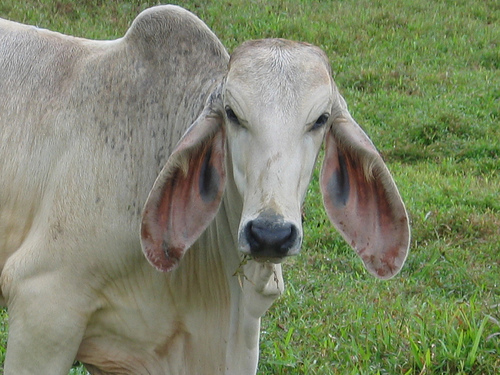} &
\includegraphics[width=\VizImageWidth,height=\VizImageHeightShort]{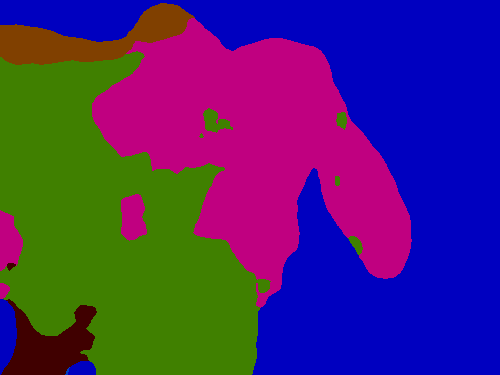} &
\includegraphics[width=\VizImageWidth,height=\VizImageHeightShort]{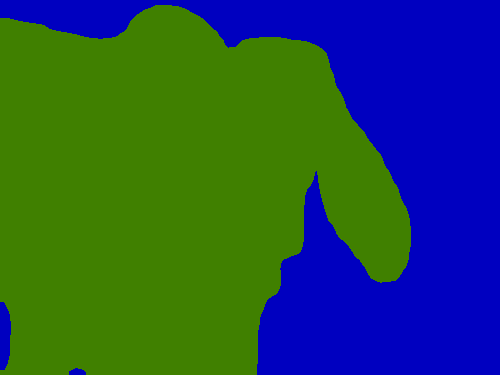} &
\includegraphics[width=\VizImageWidth,height=\VizImageHeightShort]{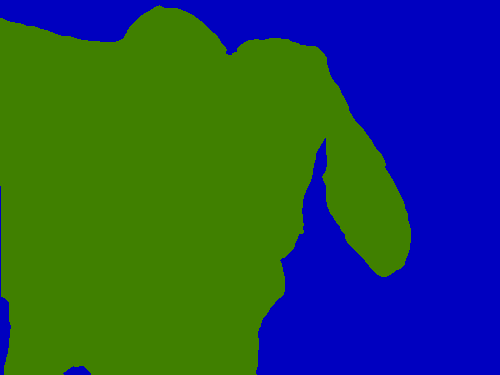} &
\includegraphics[width=\VizImageWidth,height=\VizImageHeightShort]{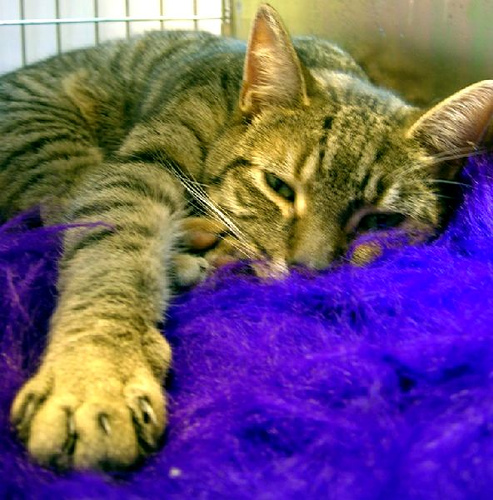} &
\includegraphics[width=\VizImageWidth,height=\VizImageHeightShort]{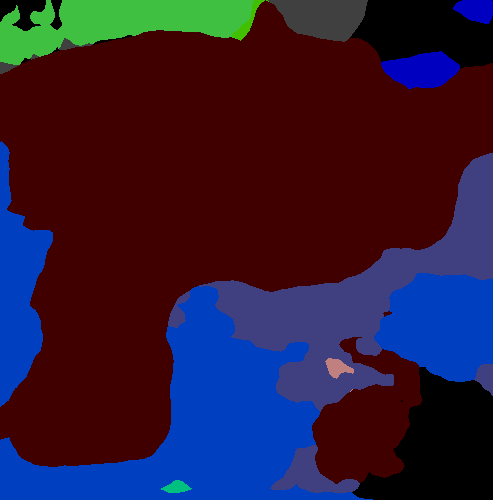} &
\includegraphics[width=\VizImageWidth,height=\VizImageHeightShort]{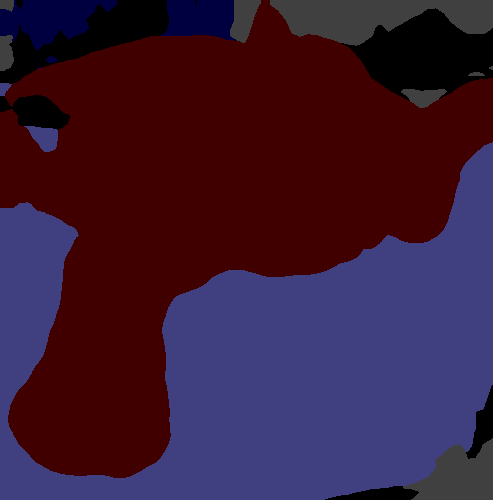} &
\includegraphics[width=\VizImageWidth,height=\VizImageHeightShort]{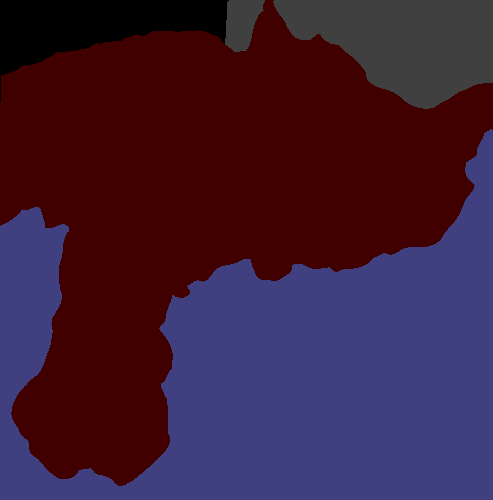} \\
\includegraphics[width=\VizImageWidth,height=\VizImageHeightShort]{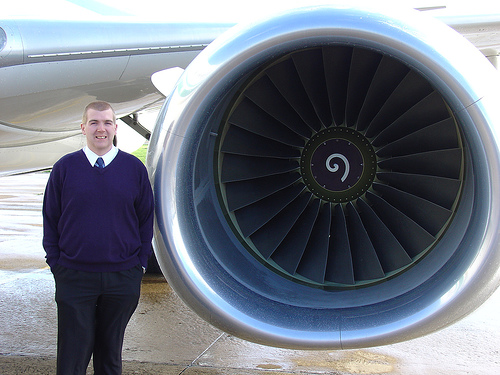} &
\includegraphics[width=\VizImageWidth,height=\VizImageHeightShort]{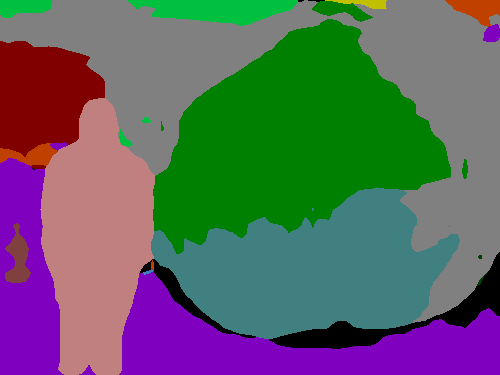} &
\includegraphics[width=\VizImageWidth,height=\VizImageHeightShort]{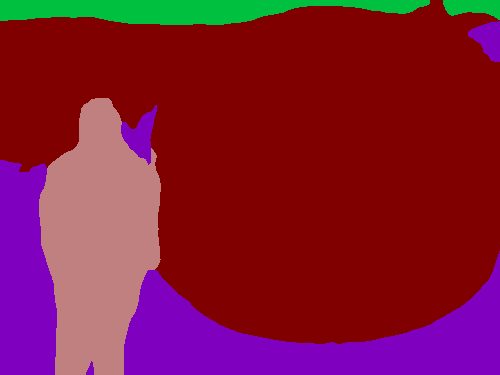} &
\includegraphics[width=\VizImageWidth,height=\VizImageHeightShort]{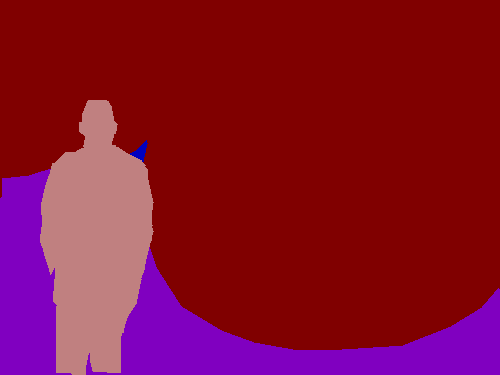} &
\includegraphics[width=\VizImageWidth,height=\VizImageHeightShort]{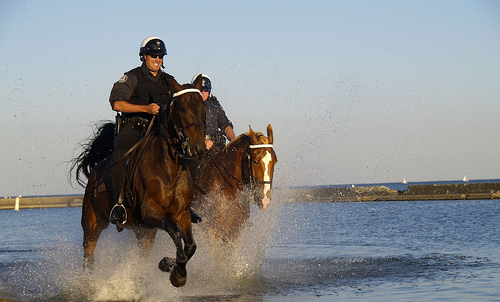} &
\includegraphics[width=\VizImageWidth,height=\VizImageHeightShort]{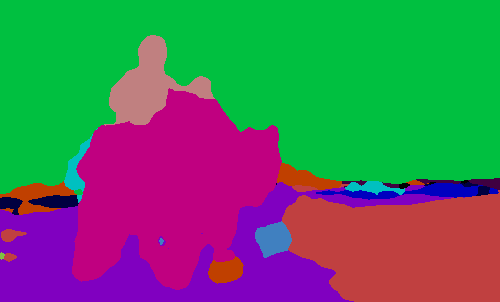} &
\includegraphics[width=\VizImageWidth,height=\VizImageHeightShort]{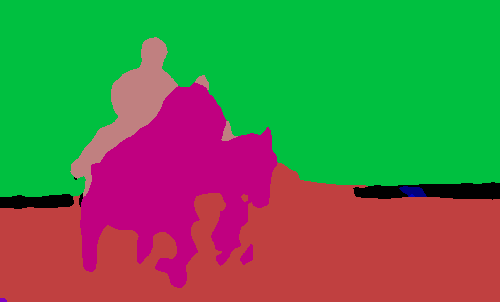} &
\includegraphics[width=\VizImageWidth,height=\VizImageHeightShort]{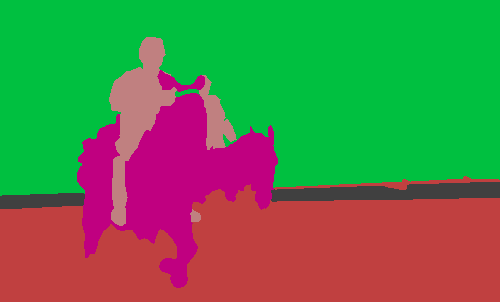} \\
\includegraphics[width=\VizImageWidth,height=\VizImageHeightShort]{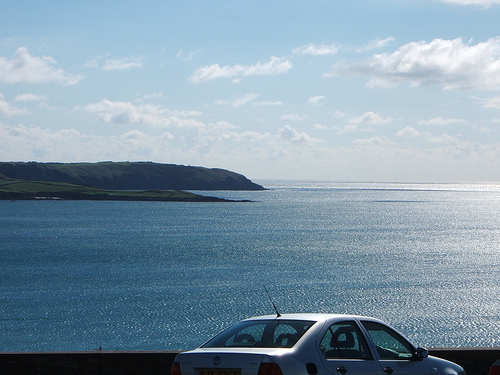} &
\includegraphics[width=\VizImageWidth,height=\VizImageHeightShort]{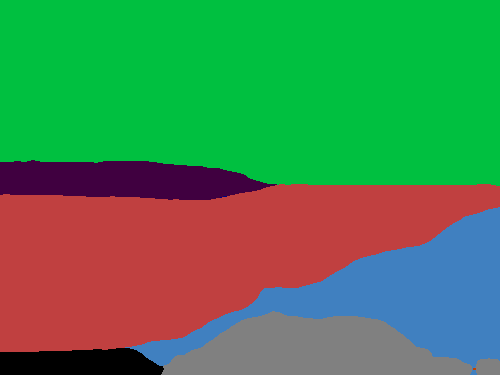} &
\includegraphics[width=\VizImageWidth,height=\VizImageHeightShort]{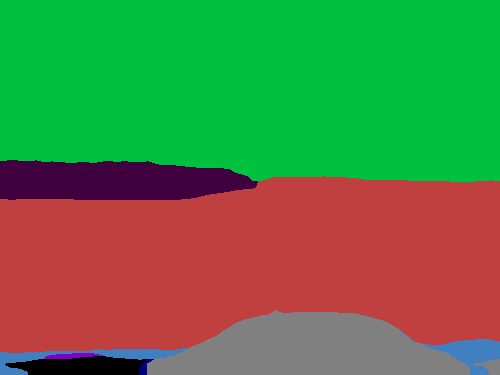} &
\includegraphics[width=\VizImageWidth,height=\VizImageHeightShort]{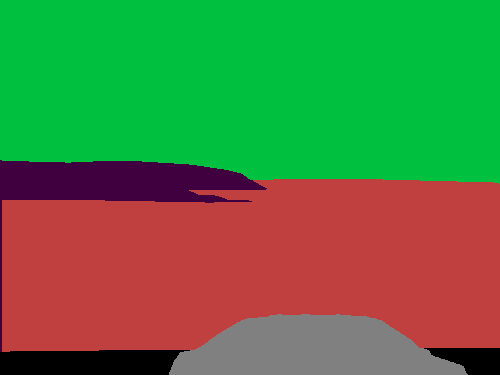} &
\includegraphics[width=\VizImageWidth,height=\VizImageHeightShort]{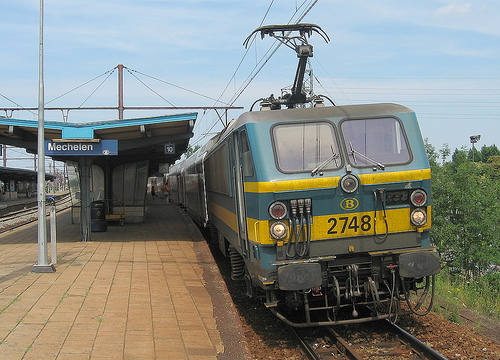} &
\includegraphics[width=\VizImageWidth,height=\VizImageHeightShort]{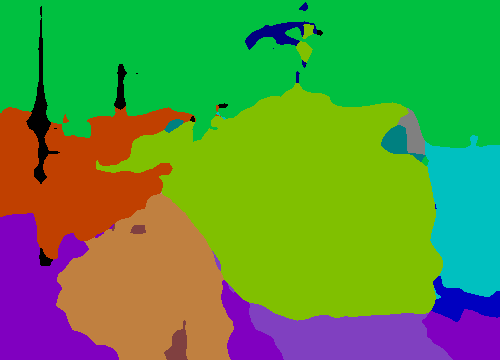} &
\includegraphics[width=\VizImageWidth,height=\VizImageHeightShort]{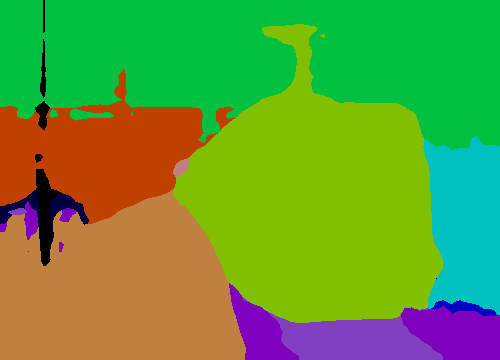} &
\includegraphics[width=\VizImageWidth,height=\VizImageHeightShort]{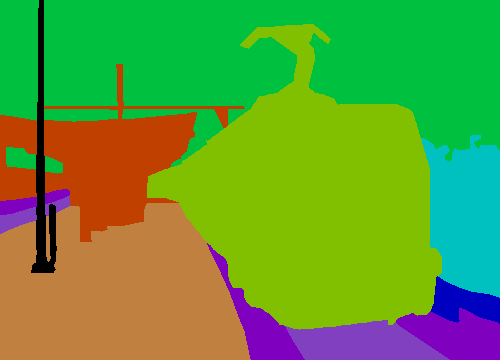} \\
\includegraphics[width=\VizImageWidth,height=\VizImageHeightShort]{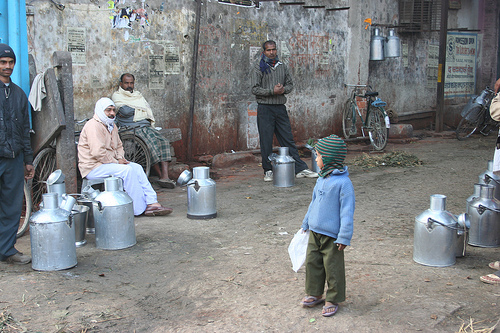} &
\includegraphics[width=\VizImageWidth,height=\VizImageHeightShort]{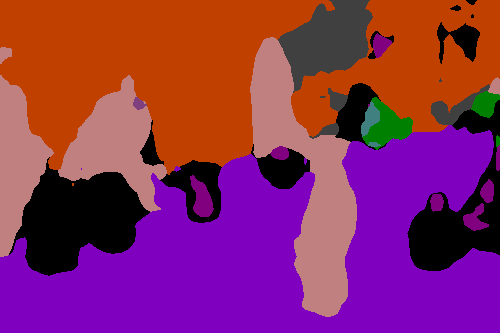} &
\includegraphics[width=\VizImageWidth,height=\VizImageHeightShort]{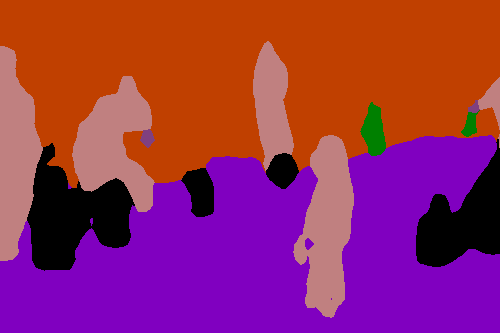} &
\includegraphics[width=\VizImageWidth,height=\VizImageHeightShort]{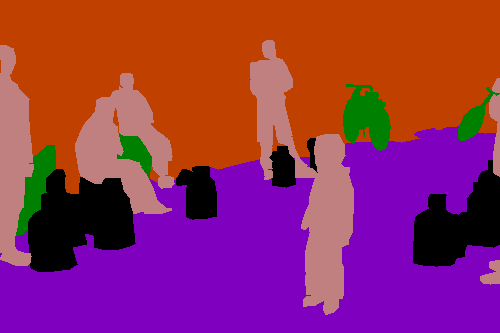} &
\includegraphics[width=\VizImageWidth,height=\VizImageHeightShort]{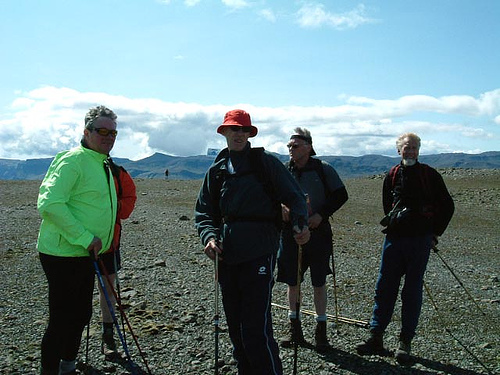} &
\includegraphics[width=\VizImageWidth,height=\VizImageHeightShort]{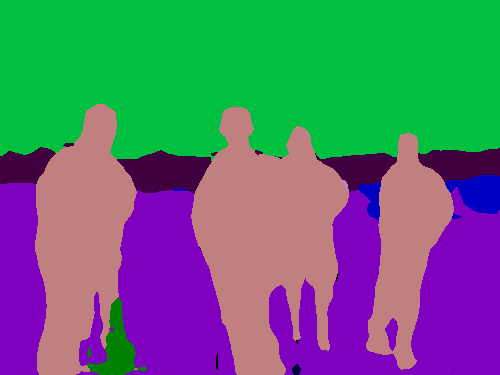} &
\includegraphics[width=\VizImageWidth,height=\VizImageHeightShort]{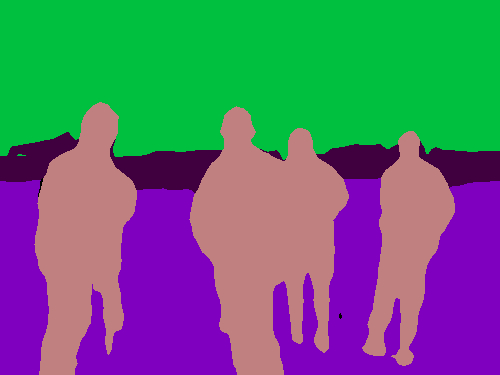} &
\includegraphics[width=\VizImageWidth,height=\VizImageHeightShort]{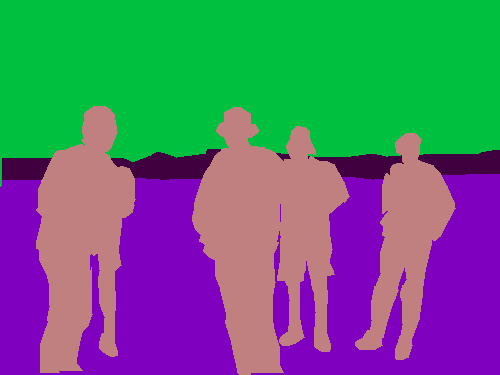} \\
\end{tabular}
\caption{Visualization of the semantic segmentation results on sample PASCAL-Context images. We compare our DC + FCN$^{+}$ with the FCN method of~\cite{shelhamer_pami16}. We also show the origin images and the ground-truth annotations for reference.}
\label{fig:sup:viz:pascal}
\end{figure*}


\begin{figure*}[tp]
\small
\centering
\tabcolsep 1pt
\begin{tabular}{cccc|cccc}
 & & DC + & & & & DC + & \\
Image & DilatedNet~\cite{zhou_arxiv16} & DilatedNet$^{+}$ & Ground Truth & Image & DilatedNet~\cite{zhou_arxiv16} & DilatedNet$^{+}$ & Ground Truth \\
\includegraphics[width=\VizImageWidth,height=\VizImageHeightLong]{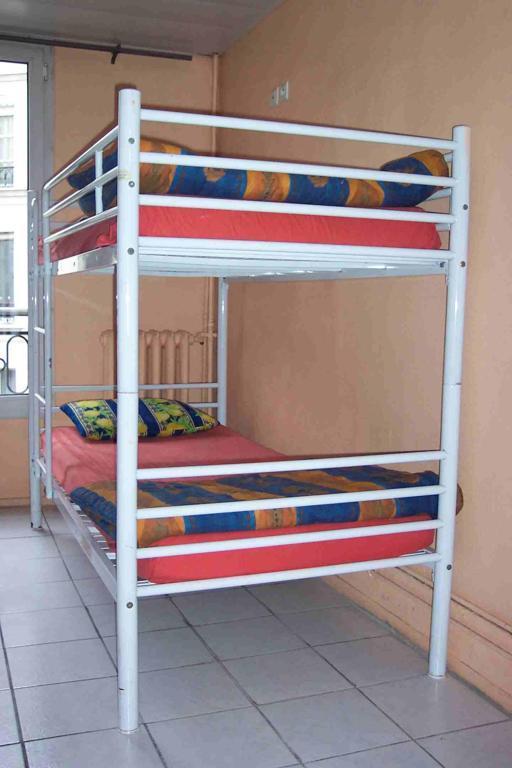} &
\includegraphics[width=\VizImageWidth,height=\VizImageHeightLong]{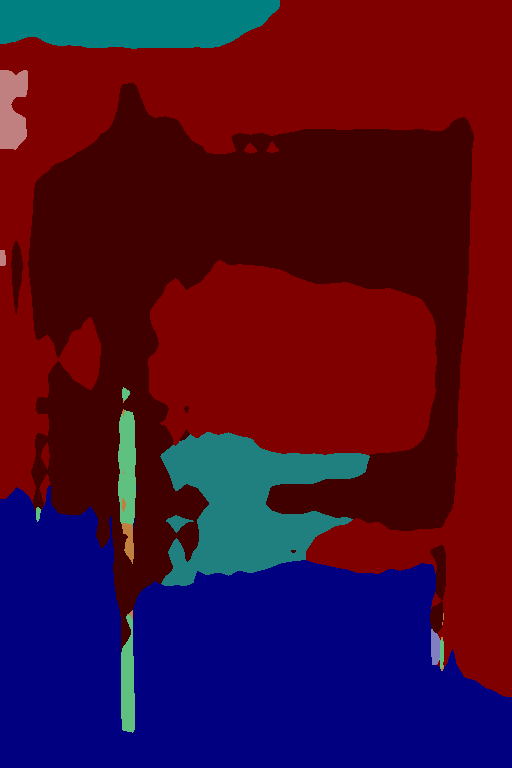} &
\includegraphics[width=\VizImageWidth,height=\VizImageHeightLong]{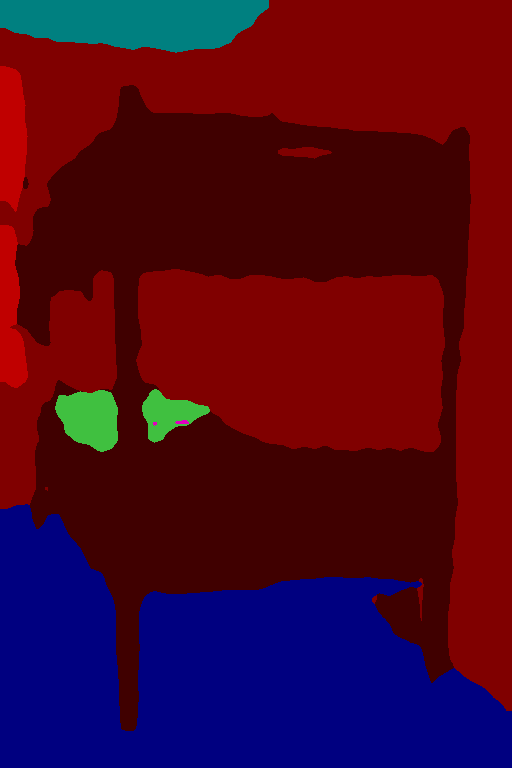} &
\includegraphics[width=\VizImageWidth,height=\VizImageHeightLong]{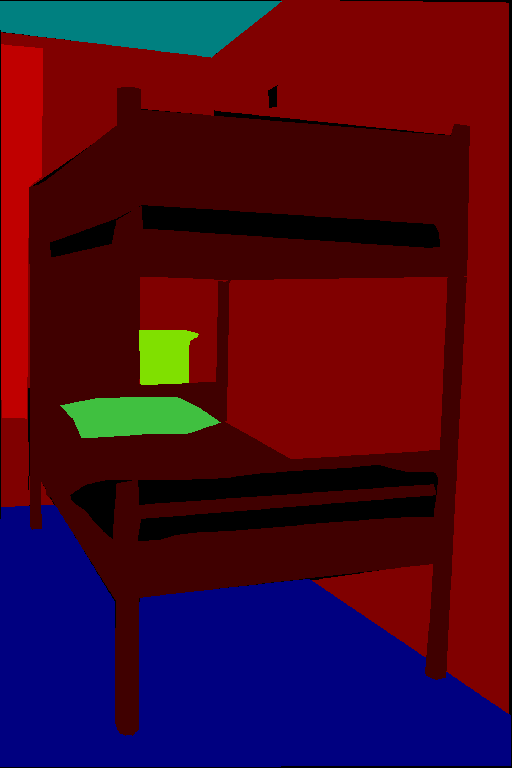} &
\includegraphics[width=\VizImageWidth,height=\VizImageHeightLong]{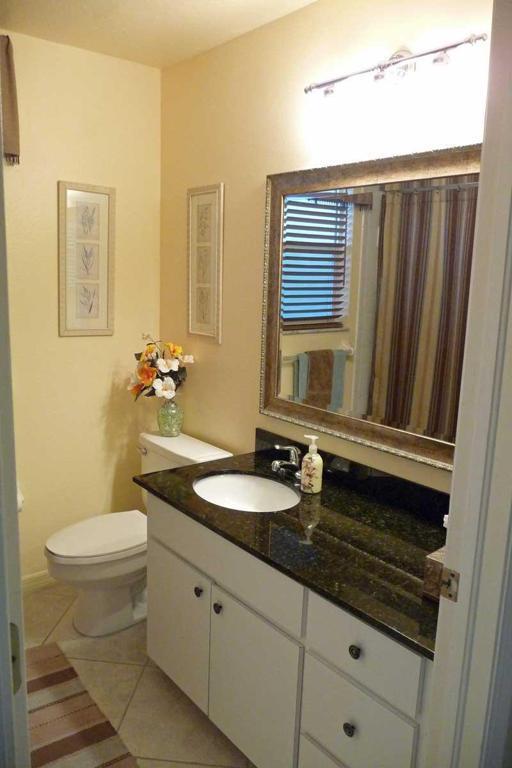} &
\includegraphics[width=\VizImageWidth,height=\VizImageHeightLong]{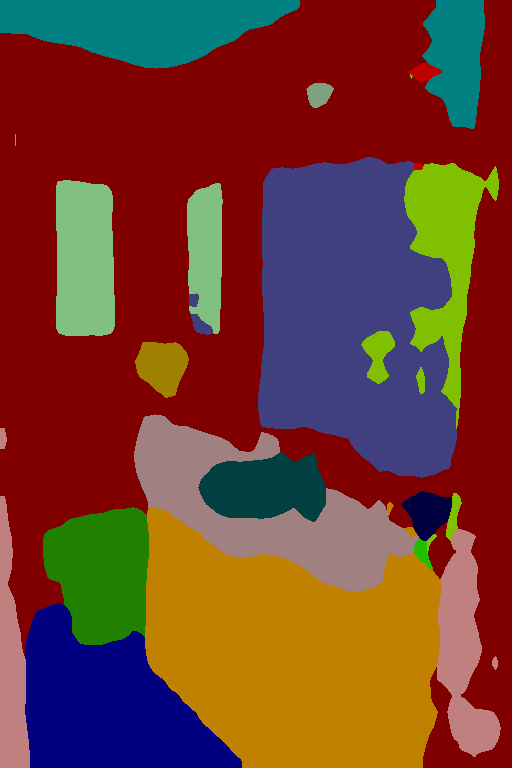} &
\includegraphics[width=\VizImageWidth,height=\VizImageHeightLong]{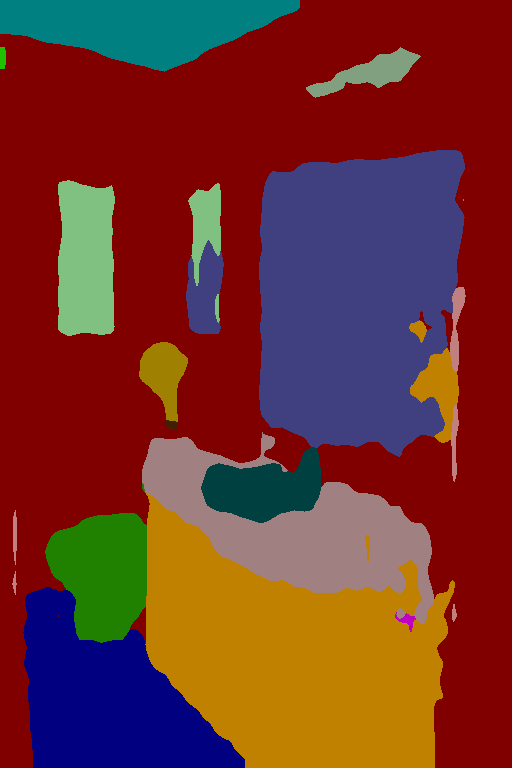} &
\includegraphics[width=\VizImageWidth,height=\VizImageHeightLong]{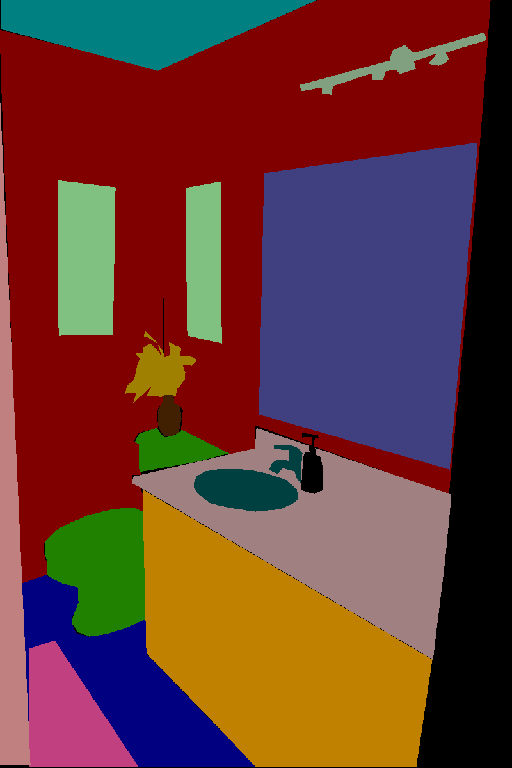} \\
\includegraphics[width=\VizImageWidth,height=\VizImageHeightLong]{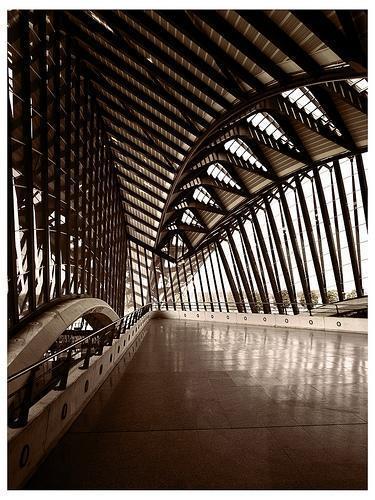} &
\includegraphics[width=\VizImageWidth,height=\VizImageHeightLong]{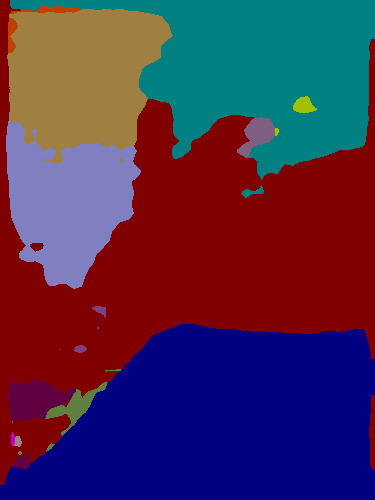} &
\includegraphics[width=\VizImageWidth,height=\VizImageHeightLong]{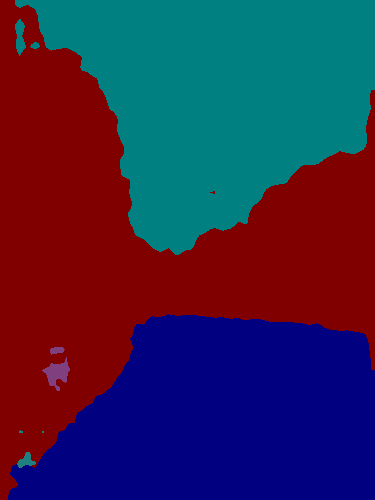} &
\includegraphics[width=\VizImageWidth,height=\VizImageHeightLong]{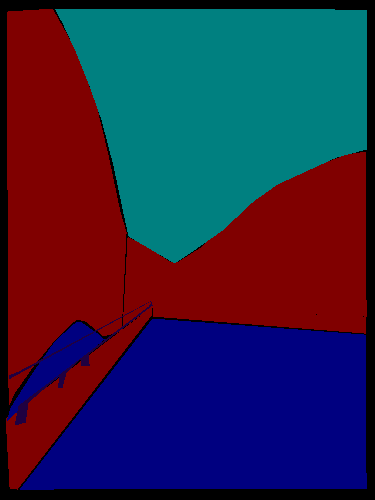} &
\includegraphics[width=\VizImageWidth,height=\VizImageHeightLong]{file/ade/ADE_val_00000028.jpg} &
\includegraphics[width=\VizImageWidth,height=\VizImageHeightLong]{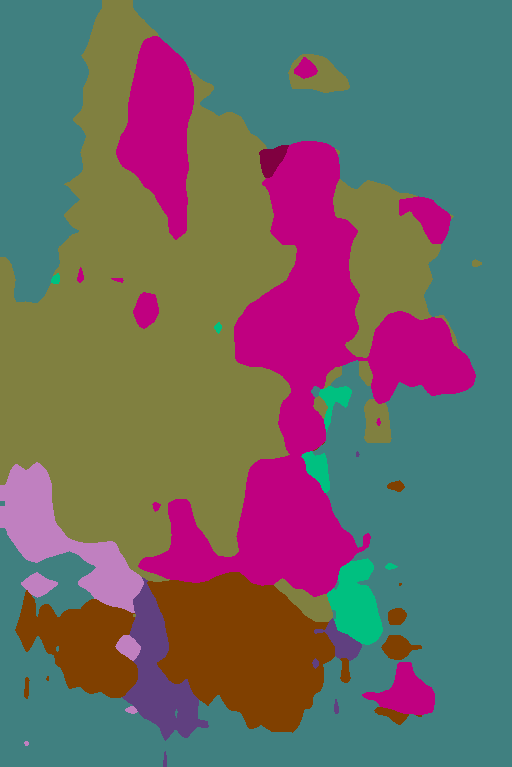} &
\includegraphics[width=\VizImageWidth,height=\VizImageHeightLong]{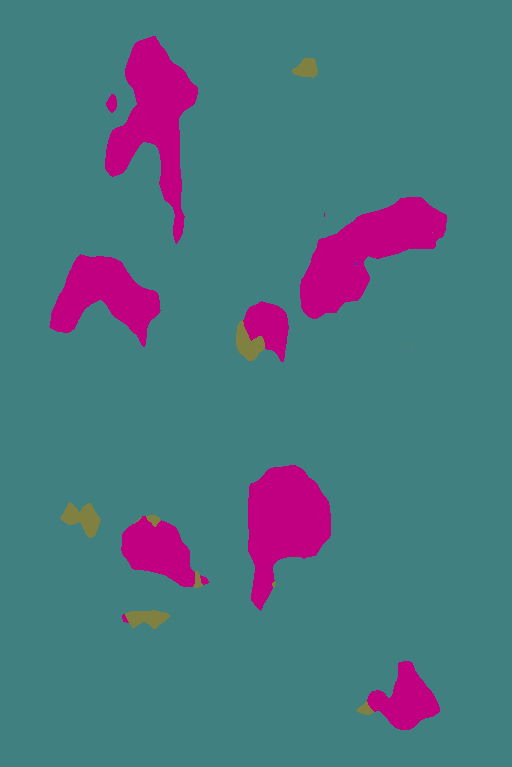} &
\includegraphics[width=\VizImageWidth,height=\VizImageHeightLong]{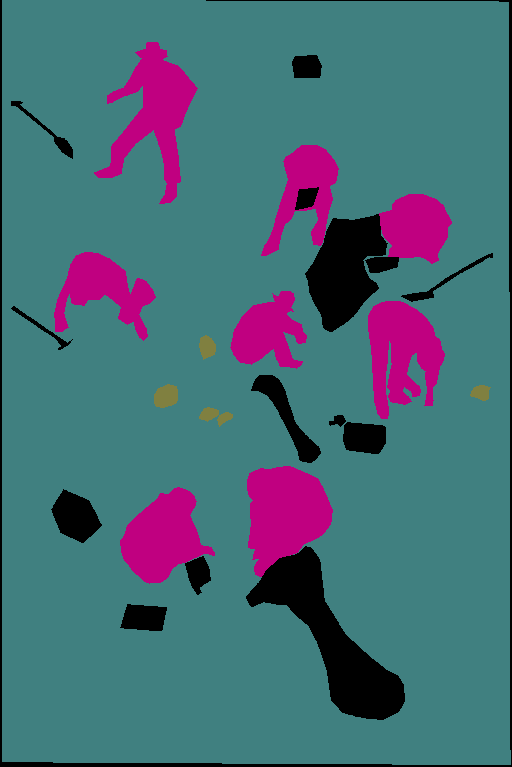} \\
\includegraphics[width=\VizImageWidth,height=\VizImageHeightMiddle]{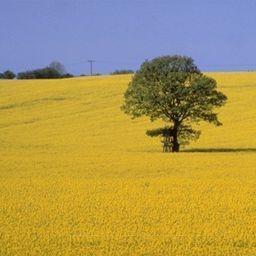} &
\includegraphics[width=\VizImageWidth,height=\VizImageHeightMiddle]{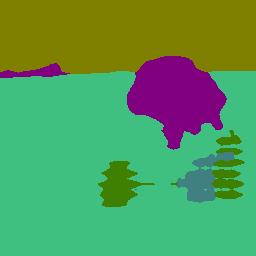} &
\includegraphics[width=\VizImageWidth,height=\VizImageHeightMiddle]{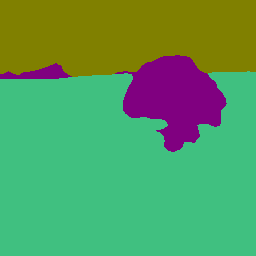} &
\includegraphics[width=\VizImageWidth,height=\VizImageHeightMiddle]{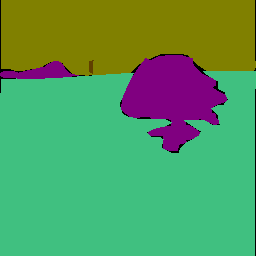} &
\includegraphics[width=\VizImageWidth,height=\VizImageHeightMiddle]{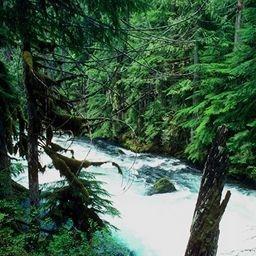} &
\includegraphics[width=\VizImageWidth,height=\VizImageHeightMiddle]{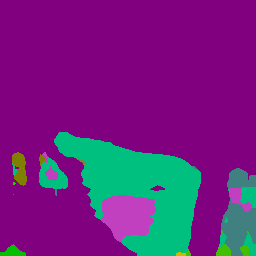} &
\includegraphics[width=\VizImageWidth,height=\VizImageHeightMiddle]{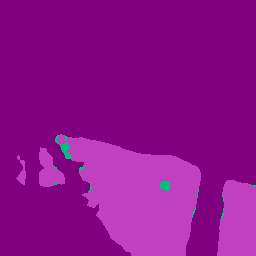} &
\includegraphics[width=\VizImageWidth,height=\VizImageHeightMiddle]{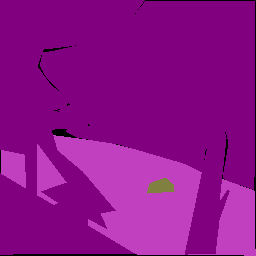} \\
\includegraphics[width=\VizImageWidth,height=\VizImageHeightMiddle]{file/ade/ADE_val_00000767.jpg} &
\includegraphics[width=\VizImageWidth,height=\VizImageHeightMiddle]{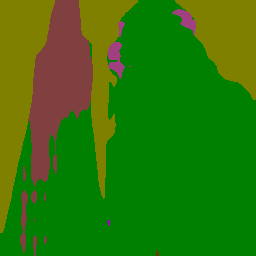} &
\includegraphics[width=\VizImageWidth,height=\VizImageHeightMiddle]{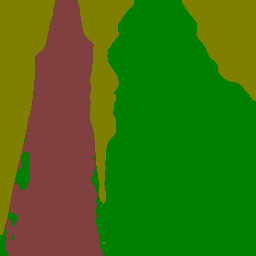} &
\includegraphics[width=\VizImageWidth,height=\VizImageHeightMiddle]{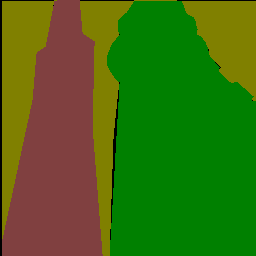} &
\includegraphics[width=\VizImageWidth,height=\VizImageHeightMiddle]{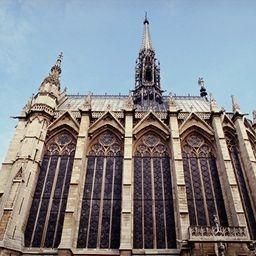} &
\includegraphics[width=\VizImageWidth,height=\VizImageHeightMiddle]{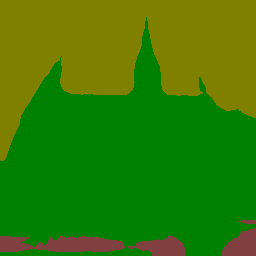} &
\includegraphics[width=\VizImageWidth,height=\VizImageHeightMiddle]{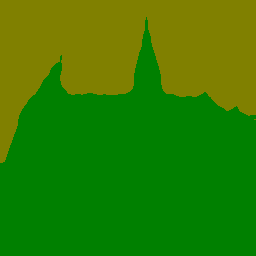} &
\includegraphics[width=\VizImageWidth,height=\VizImageHeightMiddle]{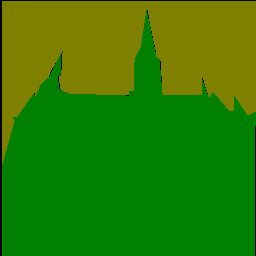} \\
\includegraphics[width=\VizImageWidth,height=\VizImageHeightShort]{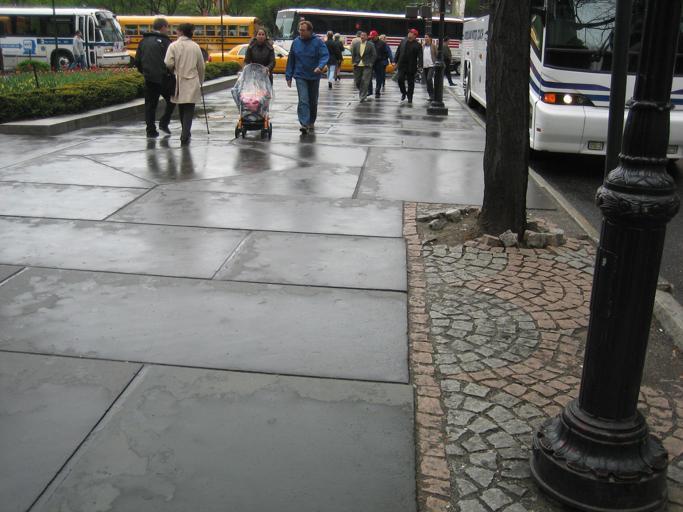} &
\includegraphics[width=\VizImageWidth,height=\VizImageHeightShort]{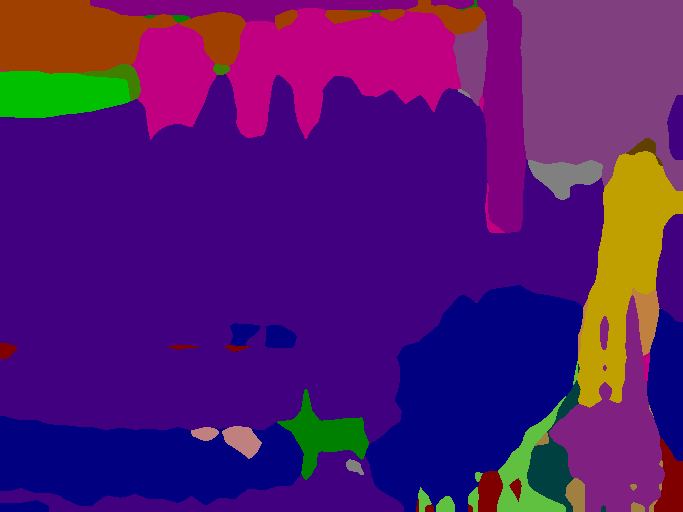} &
\includegraphics[width=\VizImageWidth,height=\VizImageHeightShort]{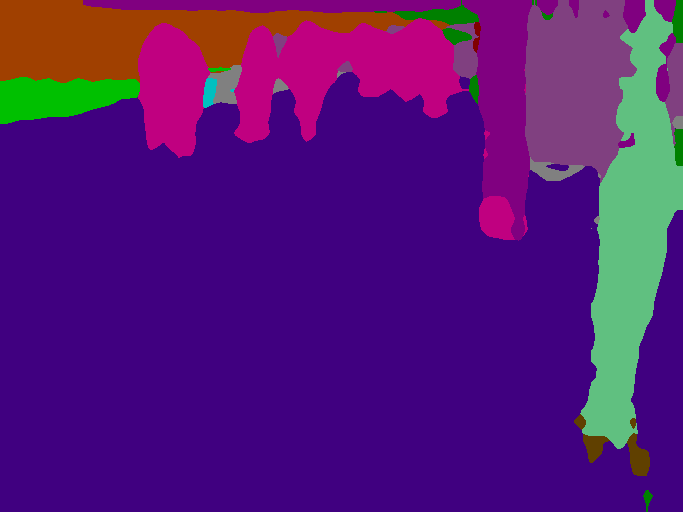} &
\includegraphics[width=\VizImageWidth,height=\VizImageHeightShort]{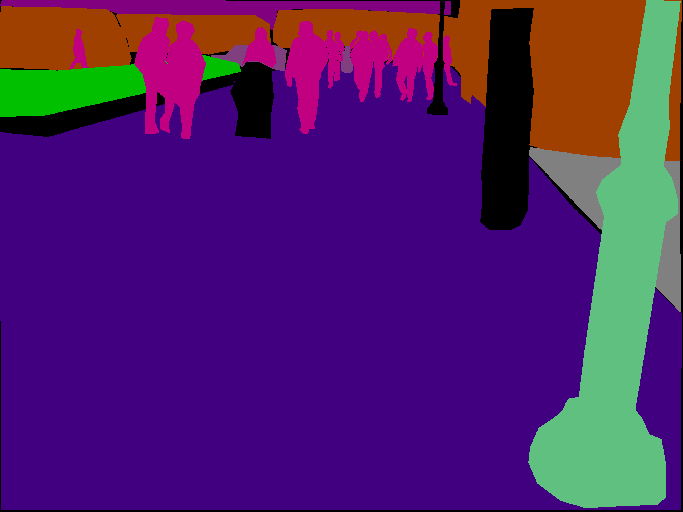} &
\includegraphics[width=\VizImageWidth,height=\VizImageHeightShort]{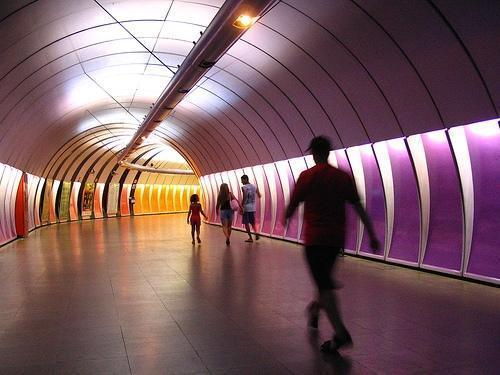} &
\includegraphics[width=\VizImageWidth,height=\VizImageHeightShort]{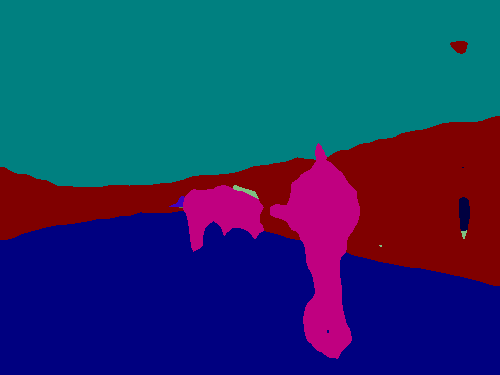} &
\includegraphics[width=\VizImageWidth,height=\VizImageHeightShort]{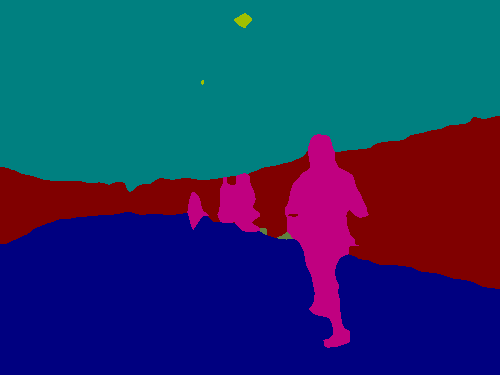} &
\includegraphics[width=\VizImageWidth,height=\VizImageHeightShort]{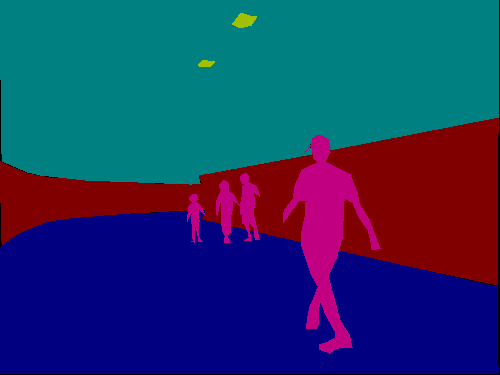} \\
\includegraphics[width=\VizImageWidth,height=\VizImageHeightShort]{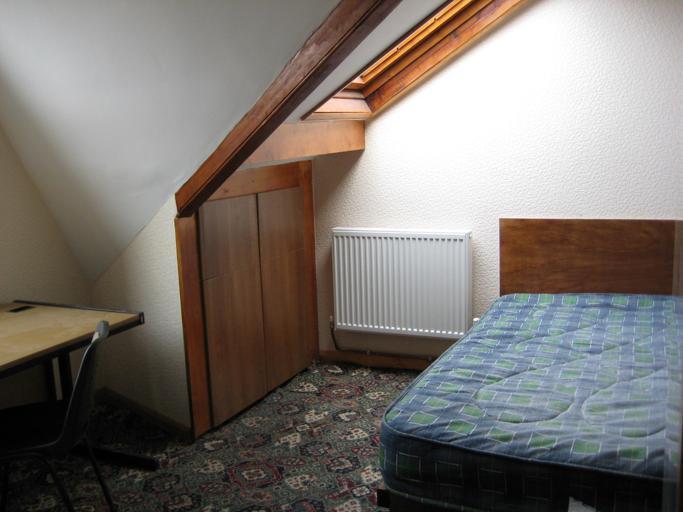} &
\includegraphics[width=\VizImageWidth,height=\VizImageHeightShort]{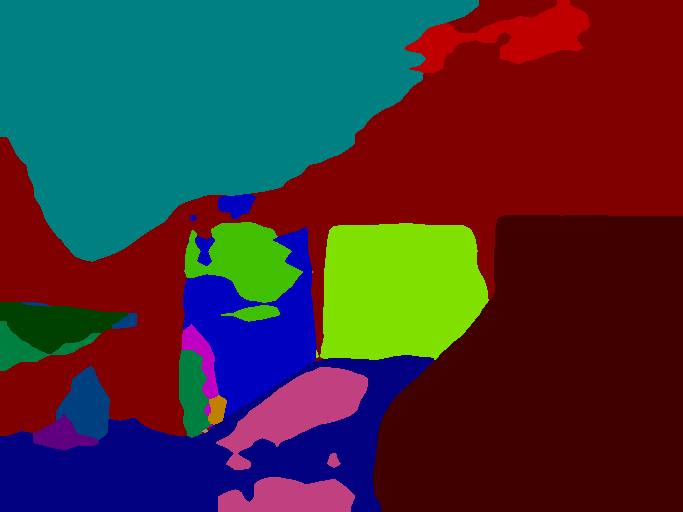} &
\includegraphics[width=\VizImageWidth,height=\VizImageHeightShort]{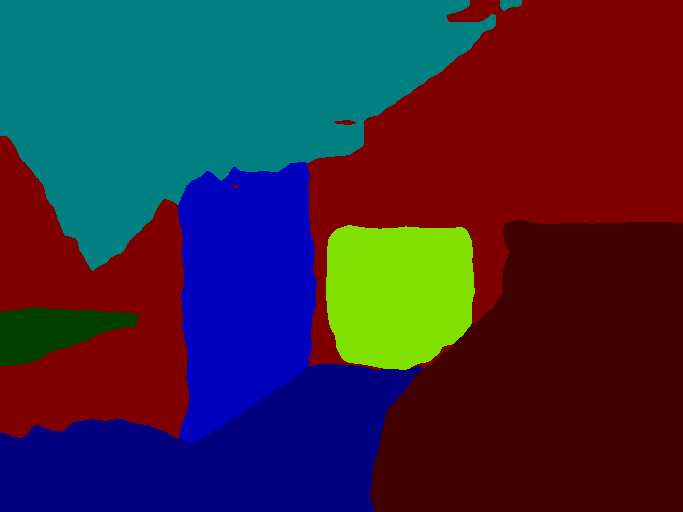} &
\includegraphics[width=\VizImageWidth,height=\VizImageHeightShort]{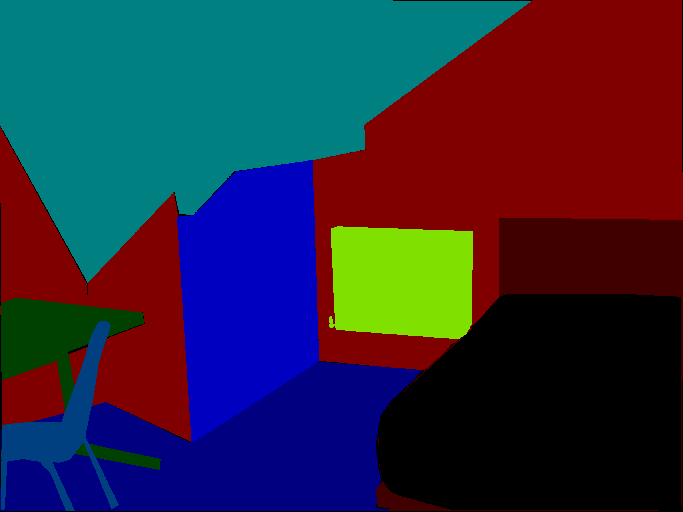} &
\includegraphics[width=\VizImageWidth,height=\VizImageHeightShort]{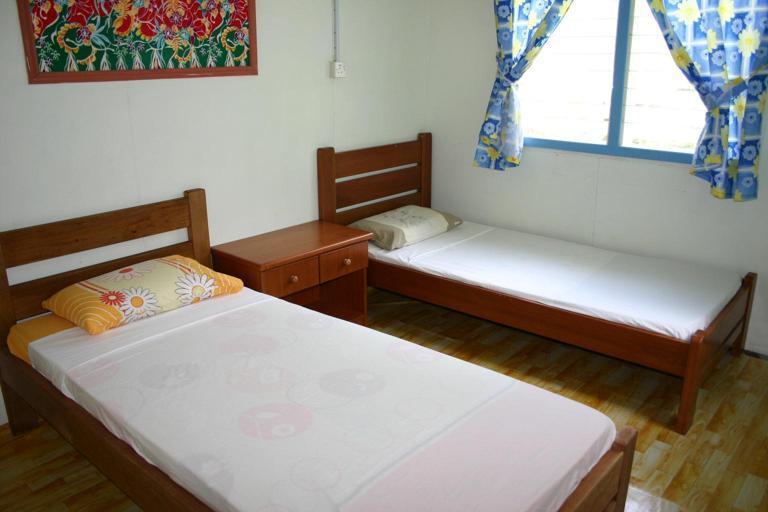} &
\includegraphics[width=\VizImageWidth,height=\VizImageHeightShort]{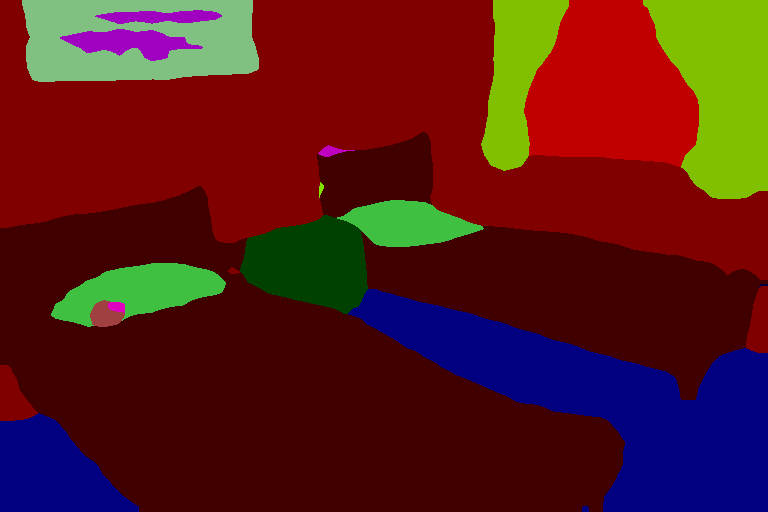} &
\includegraphics[width=\VizImageWidth,height=\VizImageHeightShort]{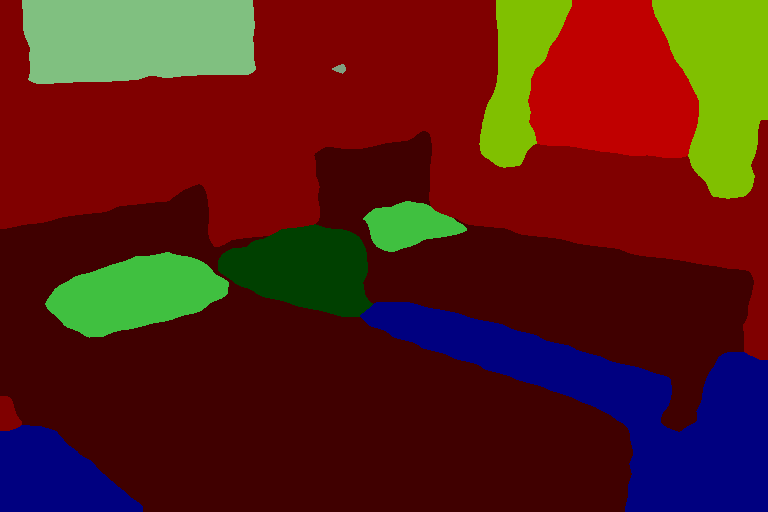} &
\includegraphics[width=\VizImageWidth,height=\VizImageHeightShort]{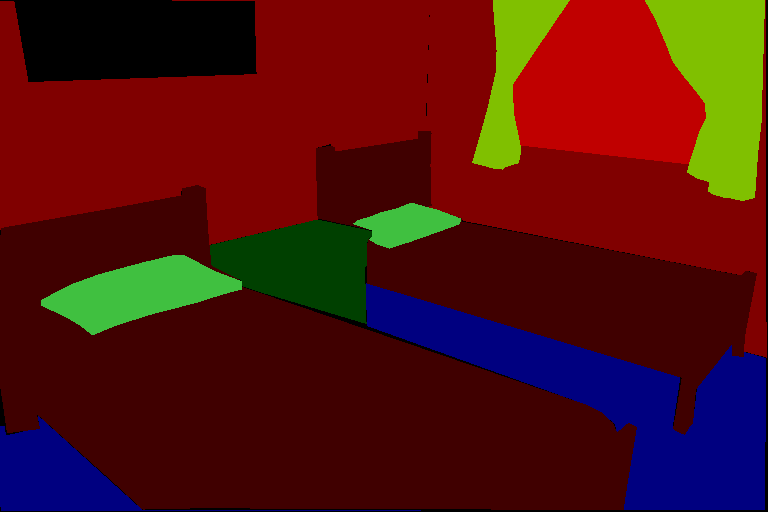} \\
\includegraphics[width=\VizImageWidth,height=\VizImageHeightShort]{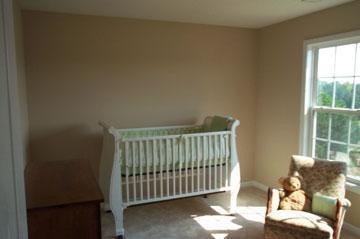} &
\includegraphics[width=\VizImageWidth,height=\VizImageHeightShort]{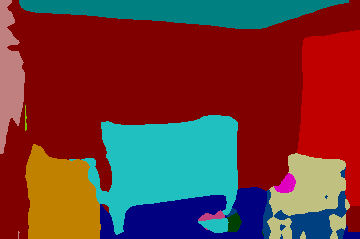} &
\includegraphics[width=\VizImageWidth,height=\VizImageHeightShort]{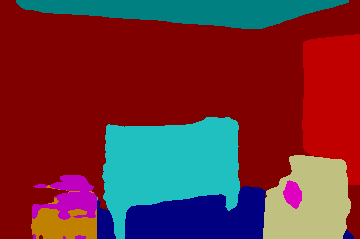} &
\includegraphics[width=\VizImageWidth,height=\VizImageHeightShort]{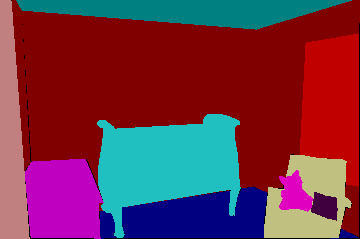} &
\includegraphics[width=\VizImageWidth,height=\VizImageHeightShort]{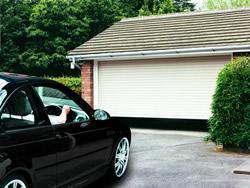} &
\includegraphics[width=\VizImageWidth,height=\VizImageHeightShort]{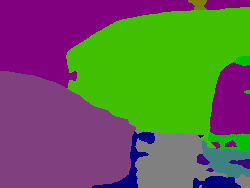} &
\includegraphics[width=\VizImageWidth,height=\VizImageHeightShort]{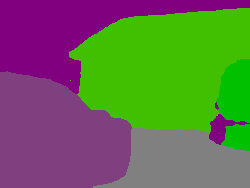} &
\includegraphics[width=\VizImageWidth,height=\VizImageHeightShort]{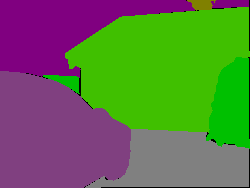} \\
\end{tabular}
\caption{Visualization of the semantic segmentation results on sample ADE20K images. We compare our DC + DilatedNet$^{+}$ with the DilatedNet method of~\cite{zhou_arxiv16}. We also show the origin images and the ground-truth annotations for reference.}
\label{fig:sup:viz:ade}
\end{figure*}

\end{appendices}

\end{document}